\documentclass[sigconf]{acmart}
\usepackage{hyperref}
\usepackage[hyphenbreaks]{breakurl}
\usepackage{balance}
\usepackage{url}
\usepackage{color}
\usepackage{caption}
\usepackage{diagbox}
\usepackage{graphicx}
\usepackage{subfig}
\usepackage{multirow}
\usepackage{booktabs}
\usepackage{epsfig,endnotes}
\usepackage{enumitem}
\newcommand{\RNum}[1]{\uppercase\expandafter{\romannumeral #1\relax}}

\newcommand{\para}[1]{{\vspace{2pt} \noindent \textbf{#1}
    \hspace{6pt}}}

\definecolor{applegreen}{rgb}{0.55, 0.71, 0.0}
\newcommand{\anna}[1]{{\color{black} #1}}
\newcommand{\annarebuttal}[1]{{\color{black} #1}}

\newcommand{\jpedit}[1]{{\color{black} #1}}

\newcommand{\htedit}[1]{{\color{black} #1}}

\newcommand{\secspace}{\vspace{-0.05in}}

\newcommand{\ad}[1]{{$\mathcal{A}$}}
\newcommand{\service}[1]{{$\mathcal{S}$}}

\newenvironment{packed_itemize}{
\begin{list}{\labelitemi}{\leftmargin=0.5em}
  \setlength{\itemsep}{1pt}
  \setlength{\parskip}{0pt}
  \setlength{\parsep}{0pt}
  \setlength{\headsep}{0pt}
  \setlength{\topskip}{0pt}
  \setlength{\topmargin}{0pt}
  \setlength{\topsep}{0pt}
  \setlength{\partopsep}{0pt}
}{\end{list}}

\newenvironment{packed_enumerate}{
\begin{enumerate}
 \setlength{\itemsep}{1pt}
 \setlength{\parskip}{0pt}
 \setlength{\parsep}{0pt}
 \setlength{\headsep}{0pt}
 \setlength{\topskip}{0pt}
 \setlength{\topmargin}{0pt}
 \setlength{\topsep}{0pt}
 \setlength{\partopsep}{0pt}
}{\end{enumerate}}

\copyrightyear{2024}
\acmYear{2024}
\setcopyright{rightsretained}
\acmConference[CCS '24] {Proceedings of the 2024 ACM SIGSAC Conference on Computer and Communications Security}{October 14--18, 2024}{Salt Lake City, UT, USA.}
\acmBooktitle{Proceedings of the 2024 ACM SIGSAC Conference on Computer and Communications Security (CCS '24), October 14--18, 2024, Salt Lake City, UT, USA}
\acmISBN{979-8-4007-0636-3/24/10}
\acmDOI{10.1145/3658644.3670306}

\begin{document}

\title{Organic or Diffused: Can We Distinguish Human Art from AI-generated Images?}

\author{Anna Yoo Jeong Ha}
\orcid{0009-0008-5551-7847}
\authornote{Both authors contributed equally to the paper}
\affiliation{
  \institution{University of Chicago}
  \city{Chicago, IL}
  \country{USA}
}
\author{Josephine Passananti}
\authornotemark[1]
\orcid{0000-0002-5705-8209}
\affiliation{
  \institution{University of Chicago}
  \city{Chicago, IL}
  \country{USA}
}
\author{Ronik Bhaskar}
\orcid{0009-0002-7524-9292}
\affiliation{
  \institution{University of Chicago}
  \city{Chicago, IL}
  \country{USA}
}
\author{Shawn Shan}
\orcid{0009-0005-4324-7817}
\affiliation{
  \institution{University of Chicago}
  \city{Chicago, IL}
  \country{USA}
}
\author{Reid Southen}
\orcid{0009-0001-4296-986X}
\affiliation{
  \institution{Concept Artist}
  \city{Detroit, MI}
  \country{USA}
}
\author{Haitao Zheng}
\orcid{0000-0002-5918-2940}
\affiliation{
  \institution{University of Chicago}
  \city{Chicago, IL}
  \country{USA}
}
\author{Ben Y. Zhao}
\orcid{0009-0003-8909-0494}
\affiliation{
  \institution{University of Chicago}
  \city{Chicago, IL}
  \country{USA}
}

\begin{abstract}
  The advent of generative AI images has completely disrupted the art
  world. Distinguishing AI generated images from human art is a challenging
  problem whose impact is growing over time. A failure to address this
  problem allows bad actors to defraud individuals paying a
  premium for human art and companies whose stated policies forbid AI
  imagery. It is also critical for content owners to establish copyright, and
  for model trainers interested in curating training data in order to avoid
  potential model collapse. 

  There are several different approaches to distinguishing human art from AI
  images, including classifiers trained by supervised learning, research
  tools targeting diffusion models, and identification by professional
  artists using their knowledge of artistic techniques. In this paper, we
  seek to understand how well these approaches can perform against today's
  modern generative models in both benign and adversarial settings. We curate
  real human art across 7 styles, generate matching images from 5 generative
  models, and apply 8 detectors (5 automated detectors and 3 different human
  groups including 180 crowdworkers, 3800+ professional artists, and 13
  expert artists experienced at detecting AI). Both Hive and expert artists
  do very well, but make mistakes in different ways (Hive is weaker against
  adversarial perturbations while Expert artists produce higher false
  positives). We believe these weaknesses will persist, and argue that a
  combination of human and automated detectors provides the best combination
  of accuracy and robustness.
\end{abstract}


\maketitle
\renewcommand{\shortauthors}{A. Y. J. Ha, J. Passananti, R. Bhaskar, S. Shan,
R. Southen, H. Zheng, and B. Y. Zhao}

\begin{figure*}[t]
    \centering 
    \includegraphics[width=1.6\columnwidth]{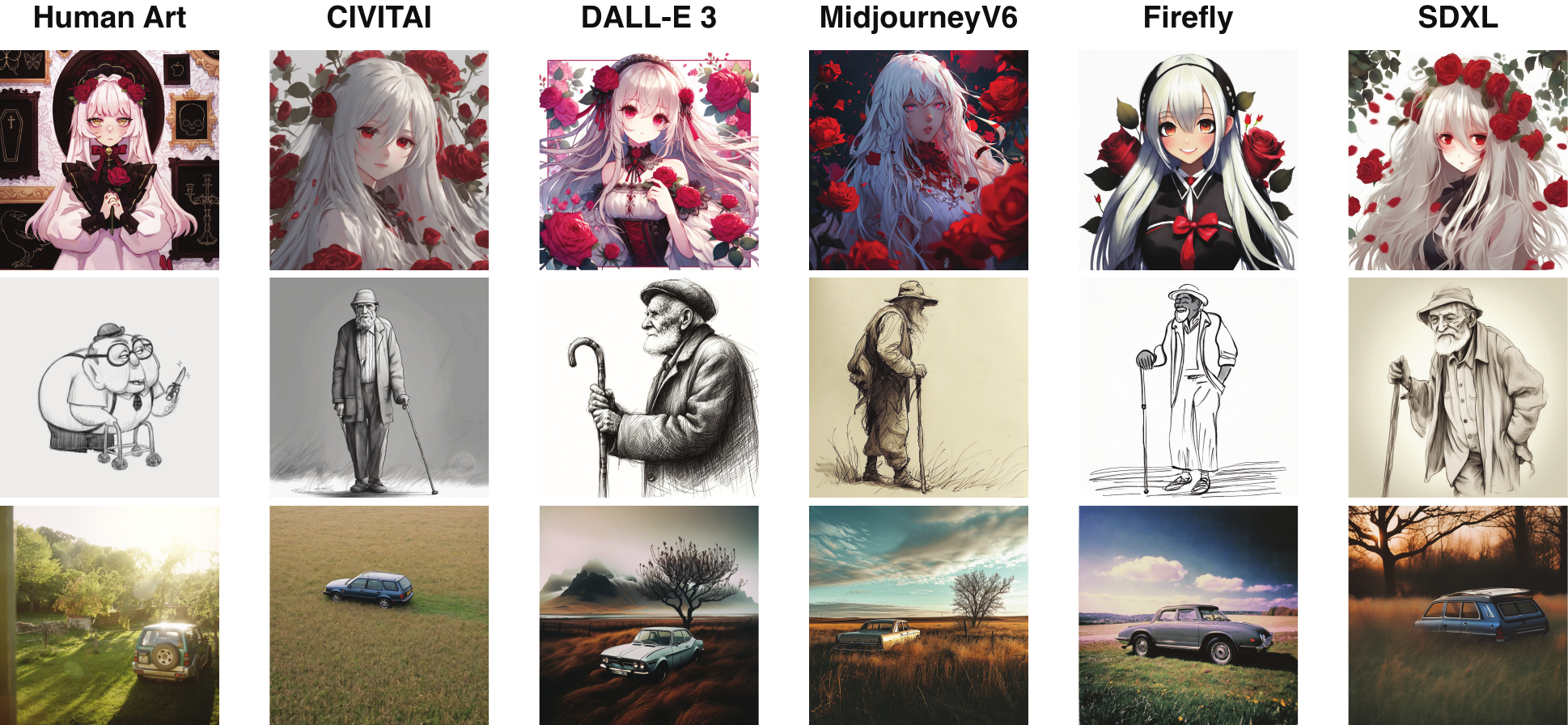}
        \vspace{-0.1in}
        \caption{Samples of human art and
          matching images produced by generative AI models. Copyright held by
          respective artists, \copyright Kirsty (@kirue\_t), \copyright
          Nguyen Viet, \copyright Liam Collod}
    \label{fig:real-5fake-img}
    \vspace{-0.1in}
  \end{figure*}

\section{Introduction}

Creative expression through artwork is intrinsic to the human
experience. From cave paintings by Neanderthals and Homo sapiens, to modern
abstract masters like Kandinsky, Pollock and Mitchell, human art connects us by
evoking shared experiences, trauma, and hopes and dreams.

Yet this might all be changing with the proliferation of output from generative image models
like Midjourney, DALL-E 3, Stable Diffusion XL (SDXL) and Adobe Firefly. Given prompts
as short as a single word, these models can generate glossy images that at a
glance, resemble the work of a professional artist or fine photographer. As they
continue to evolve, it is becoming increasingly difficult to distinguish art
produced by human creatives and images produced by generative AI.

Identifying if a piece of art is human-made or AI-generated is critical for a number
of reasons. First, individuals and companies are often willing to pay a
premium for human art over AI content. Also, companies or creative groups
have policies restricting the use of AI-generated imagery in competitions,
work product, or ad campaigns. Yet recent news is littered with examples of
fraud, where AI-generated images are sold as human art to
individuals~\cite{fakeart-indi,fakeart-daphne} and
publishers~\cite{fakeart-eons,fakeart-dnd}, and used in ad campaigns or
submitted to creative competitions against AI
policies~\cite{fakeart-contest,kickstarterai-policy}. This has resulted in numerous
controversies~\cite{fakeart-backlash,fakeart-backlash2,wotc-sorry}, retracted ads and
publications~\cite{riotlol-arttakedown, fakeart-dnd2}, and public
apologies~\cite{maxon-sorry,fakeart-eons}.

Second, identification of AI images is also a legal and regulatory
issue. Commercial companies want to copyright their creative content, but the
US Copyright Office has ruled that only human created artwork (or human
contributions to hybrid artwork) can be
copyrighted~\cite{fakeart-copyright}. Thus businesses using generative AI
might try to pass off AI images as human art to obtain copyright. Finally,
multiple projects have shown evidence that both text and image AI models will
degrade if only trained on output of AI
models~\cite{shumailov2023curse,genai-mad2,genai-mad}. Thus AI model trainers
also need to distinguish AI-generated images from human art for training
purposes.

All of this begs the question, do we have the tools today to reliably and
consistently distinguish AI-generated imagery from human-created art? There
are multiple potential solutions. First, human artists are often quite good at
recognizing human art, and some experts have demonstrated a
consistent ability to detect AI images trying to pass as human
art~\cite{fakeart-amazon,fakeart-daphne,fakeart-contest}. Alternatively,
specific companies like Hive, Optic and Illuminarty have 
trained supervised classifiers to distinguish AI imagery from human
art. Online media has raised questions on the accuracy of these detectors
amid their growing impact on news media~\cite{detectors-abuse}.
Finally, recent research results like DIRE and
DE-FAKE~\cite{Wang_2023_ICCV,sha2023defake} suggest specific techniques
to recognize images produced by diffusion models, including all of the major
generative models today. 

The goal of our work is to systematically and comprehensively explore the
effectiveness of these detectors at distinguishing AI-generated images and human
art\footnote{Unlike most prior studies, our focus is not identifying
deepfake images from real photographs. Instead our focus is on determining
the provenance of creative art imagery, and our inclusion of photography focuses
on photographs as art, not as records of real events.}.
We consider a wide range of imagery, including 7 distinct art styles,
each represented by samples of human art,  images generated by each
of 5 generative AI models, AI images painted over by humans (hybrid), and
human photography enhanced by AI. We also consider a range of ``detection
methods,'' including 5 automated tools (3 deployed classifiers Hive, Optic,
Illuminarty and 2 research detectors DIRE and DE-FAKE) and 3 different
populations of ``human detectors'' (crowdsourced non-artists, professional
artists, and expert artists). Finally, we also consider adversarial
scenarios, where AI-generated images are augmented with noise and adversarial
perturbations with the intention of bypassing detection.

In total, we curated a dataset of 280 real human art images across 7
different styles, and 350 generated AI-based
images from generative models using prompts automatically extracted from
each of the human art images. One component of our study tests the efficacy of
automated detectors on these images and their perturbed variants; and the other
part evaluates human-based 
detection. \htedit{The latter} involves 3 separate user studies on: a) 180 crowdworkers on
the Prolific platform, b) 3800+ professional artist volunteers 
\htedit{recruited} from social media artist groups, and c) 13 expert professional artists who
have experience identifying generative AI images.

Our study produces a number of significant findings:
\begin{packed_itemize}
  \item We find that normal, non-artist users are generally unable to tell
    the difference between human art and AI-generated images. Professional
    artists are more confident and perform much better on average, and expert
    artists more so.
  \item Supervised classification does surprisingly well, and Hive
    outperforms all detectors (human and ML), and produces zero false
    positives. Unsurprisingly, accuracy seems to correlate with expected
    training data availability: biggest classifier (Hive) performs best; all
    classifiers perform the worst on Firefly, the newest of the major
    generative models.
  \item Fine tuned models that mimic human artists, and real images modified
    with AI upscaling posed no significant challenges to human or ML
    detectors. 
  \item Adversarial perturbations did have significant impact on ML detectors
    such as Hive, with feature space perturbations being the most effective.
  \item Expert human artists perform very well in detecting AI-generated
    images, but often attribute mistakes and lower skill by human artists as
    evidence of AI images, thus producing false positives.
    \item  \htedit{A combined team of human and automated detectors provides the best combination of accuracy and robustness.}
\end{packed_itemize}

\secspace
\section{Background and Related Work}
\label{sec:back}
We begin by providing background on generative AI image models using the
diffusion model architecture, and on currently available automated detectors
of generative AI images. We then discuss existing work related to our study.

\secspace
\subsection{Generative Image Diffusion models}
First introduced in 2020, diffusion models quickly replaced GANs as the state-of-the-art models for image
synthesis~\cite{dhariwal2021diffusion}. This was quickly followed by
extension to text-to-image generation~\cite{ramesh2022hierarchical,
nichol2022hierarchical}.
Later instances included multiple open-source models from Stable
Diffusion~\cite{rombach2022high, sd-release, podell2023sdxl, sd-14-modelcard,
sd-15-modelcard}, and commercial models from Midjourney,
DALL-E, and Adobe Firefly.

Diffusion models face a growing number of social and ethical concerns. Base
models require enormous amounts of training data, often obtained without
consent through web-scraping. Midjourney trained their model using data from
over 16,000 artists, the vast majority without consent~\cite{mid-list}.
Stability AI trained on datasets from
LAION~\cite{schuhmann2022laion,rombach2022high}, containing millions of
copyrighted works. These copyright infringement issues have led to multiple
class-action lawsuits~\cite{sdlitigation, class-action}. Even in smaller
volumes, artists are finding their works being used without consent in finetuned
models using techniques such as LoRA~\cite{hollie-steal}.

As these models continue to improve in quality, many of their users have attempted to pass
AI-generated images as human art. AI-generated images have been used to win
art competitions, fooling judges of digital art, photography, and book
covers~\cite{winaward, photographycomp,fakeart-contest}. Companies
that promote human art have found themselves using AI content provided to
them by third-party vendors~\cite{magicthegatheringcontroversy}.

\secspace
\subsection{Automated AI-Image Detectors}
A number of software and web services offer the ability to detect if an image
is generated by generative AI image models. We group these detectors into two
categories: deployed commercial detectors and research-based detectors.

\para{Research-based detectors.} These come from published research papers that often
offer source code and training/testing datasets, sometimes also with pretrained
models. While they lack the same public reach and
influence as commercial detectors, they offer the benefit of transparency in
methodology. Two such detectors are the DIRE detector~\cite{Wang_2023_ICCV}
and DE-FAKE~\cite{sha2023defake}. DIRE, or Diffusion Reconstruction Error,
aims to exploit the forwards and backwards diffusion
processes to identify images generated by that
model. DE-FAKE uses both image-only and multimodal
embeddings~\cite{radford2021learning} to create a model-agnostic
detector~\cite{sha2023defake}. 

\para{Commercial detectors.} These detectors are deployed online, generally as
web services
with a tiered pricing model and a web-based non-transparent (black-box)
detection API. They provide easy access to image classification with minimal
computational requirements, Among the most popular are Hive AI Detector
(Hive), Optic AI or Not (Optic), and Illuminarty~\cite{hivedetection,
aiornot, illuminarty}. All three services advertise free demo plans via a web UI
with high accuracy, and are well covered by popular media~\cite{newsstory1,
newsstory2, newsstory3}.

Beyond DIRE and DE-FAKE, other techniques have been proposed to detect
AI-generated images.
Diffusion models can embed invisible watermarks
into images during generation~\cite{braudaway1997protecting}.
Diffusion watermarks involve manipulating the initial noise vector, creating
watermarks robust to some perturbations~\cite{wen2023treerings,
zhang2024robust}. However, this approach requires modifying the diffusion
model. Finally, other detection methods make use of frequency domain analysis
to detect AI-generated images as outliers~\cite{bi2023detecting}. 

\secspace
\subsection{Related work}

\para{Detecting Deepfake Photographs.} While our study focuses on
distinguishing human art from AI-generated images, several prior studies have
focused on human detection of deepfake photos generated by machine learning
models. \cite{bray2023testing} evaluates users' ability to detect deepfakes
of human faces using StyleGAN2~\cite{karras2020analyzing}, and finds
that human participants have below 65\% accuracy in all experiments, even
when taught how to recognize deepfakes. Similarly, \cite{lu2023seeing}
evaluates human detection on a set of real photos and photorealistic
images from Midjourney V5. They also create a dataset of
roughly two million fake images to train ML detectors. While humans
misclassify 37\% of images, the best-performing ML detector misclassifies
13\% of the same evaluation set.

Several projects explore robust evaluation and robust training techniques to
improve detection accuracy. \cite{wang2020cnngenerated} proposes training
data augmentation using flipping, blurring, and JPEG compression;
\cite{song2023robustness} evaluates detection under perturbations of color
contrast, color saturation, blurring, and pixelation;
\cite{bammey2023synthbuster} performs data augmentation with JPEG
compression; and \cite{hooda2024d4} uses an ensemble of detectors over the
frequency domain to improve detection robustness.

\para{Explainability in Image Identification.} Some have explored
explainability in detecting AI images. \cite{ricker2023towards}
studies distributions of GAN, diffusion, and real images, showing greater
overlap between diffusion and real distributions than between GANs and real
distributions. \cite{bird2023cifake} creates a counterfactual, generated
dataset to CIFAR-10~\cite{krizhevsky2009cifar} and uses gradient heatmaps to
visualize important features for detection. \cite{corvi2023detection}
performs forensic analysis on the frequency domain distributions of various
diffusion and GAN models.

\para{AI Images and Art.} The abundance of prior work has almost entirely
focused on detecting deepfakes and photorealistic images, including some very
large fake image benchmarks~\cite{zhu2023genimage, lu2023seeing}.
DE-FAKE briefly mentions detecting art but only tests on 50 pieces of human
art and 50 AI-generated images. Deepart~\cite{wang2023benchmarking} is an
art-based dataset composed of a random selection of images
selected from LAION-5B~\cite{schuhmann2022laion}, designed as a training
dataset for a classifier to detect AI-generated art.

The most related work on this topic was presented recently at IEEE S\&P
2024~\cite{cispa-detect}. Where our work focuses entirely on creative visual
art, this prior study covered generative AI detection
broadly across images, audio and text across Internet users in multiple
countries. 
\secspace
\section{Methodology}
\label{sec:setup}

AI-generated images have already become exceptionally good at mimicking human
art. Distinguishing these generated images from human art is critical
for individual and institutional consumers, for copyright reasons, and for AI
models seeking to curate their training datasets. The goal of this study is
to understand how feasible this task is today given recent advances in
these generative models, how and why current detectors make mistakes, and
what that portends for the future.

\secspace
\subsection{Overview: Goals and Challenges}
In tackling this multifaceted problem, our goal is to try to explore several broad
questions on this topic:
\begin{packed_enumerate}
  \item Are there detection methods today, human or automated, that can
    accurately distinguish between human art and AI-generated images? How do
    artists using their knowledge of art fundamentals fare against
    semantically-agnostic supervised classification and research tools
    designed specifically to detect diffusion model output?
  \item What are limitations of current detectors, and why do they make mistakes?
  \item How well do detectors perform under adversarial conditions,
    i.e. against images altered to avoid identification?
  \item Are there fundamental trends in performance of detection approaches,
    and what are implications as models continue to evolve? \vspace{-0.03in}
\end{packed_enumerate}
As the first research study to perform a comprehensive analysis of
classifying human art and AI-generated images, our most significant challenge
is how to capture the numerous dimensions of this problem. Most
specifically, we consider and incorporate five different dimensions into our
study. We summarize these here and present further details as we describe our
experimental methodology in the remainder of this section.
\begin{packed_itemize}
\item{\bf Art Styles.} Generative AI image models have a wide range of
success when mimicking different styles of art. Therefore, our evaluation
must cover a wide range of art styles, from anime to sketches to fine
photography.
\item{\bf Sources/Types of AI-Generated Images.} Different AI models
vary in their ability to mimic human art. Thus we must consider a
representative set of current diffusion models, as well as more
unorthodox image types such as hybrid (AI-generated images
painted/altered by humans) and upscaled (human-generated photography
expanded in resolution using AI models).
\item{\bf Range of Automated AI Image Detectors.} We include results of the
most popular available automated detectors (Hive, Optic, Illuminarty) as
well as research prototypes (DIRE, DE-FAKE).
\item{\bf Range of Human AI Image Detectors.} Humans will vary
significantly in their ability to identify human art vs AI images,
depending on their knowledge and experience in producing art. We consider
three user groups: regular users (non-artists), professional artists, and
expert artists experienced in identifying AI images.
\item{\bf Range of Adversarial Countermeasures.} Multiple factors
incentivize AI model users to alter their images to escape identification
as AI images. Thus our study also considers multiple types of adversarial
perturbations and explore their ability to confuse different detectors
(both automated and human).
\end{packed_itemize}

\subsection{Evaluating Automated Detectors}

For automated software-based detectors, we consider both deployed commercial
systems, as well as research-based systems. There are three well-known deployed
commercial systems:

\begin{packed_itemize}
  \item Hive: AI content detection using supervised classification provided by thehive.ai.
  \item Optic: ``AI or Not'' is a free service (for limited queries) running a proprietary
  algorithm to detect AI images and audio.
  \item Illuminarty: an AI detection service running a proprietary
  algorithm including an implementation of DIRE.
\end{packed_itemize}
  
For research-based systems, we selected two recent systems that had code (and models)
available for testing.
\begin{packed_itemize}
  \item DIRE~\cite{Wang_2023_ICCV}: DIRE (Diffusion Reconstruction Error)
  pushes a test sample forwards and backwards through a diffusion
  pipeline and measures its changes to detect if the image came from that
  pipeline. DIRE has pretrained models with a public implementation.
  \item DE-FAKE~\cite{sha2023defake}: DE-FAKE uses both image-only and
  multimodal embeddings to create a model-agnostic detector. We trained a
  model based on techniques from the paper.
\end{packed_itemize}

We evaluate both automated detectors and human detection on the same core
test dataset of images (280 human art pieces, 350 AI images, 40 hybrid
images), described in more detail in Section~\ref{sec:dataset}.
However, we also test automated detectors against a variety of adversarial
perturbations including Gaussian noise, JPEG compression, adversarial
perturbations, and the Glaze style mimicry protection tool~\cite{shan2023glaze}.

\subsection{Evaluating Human Detection: User Studies}
\label{subsec:humansetup}

Recent events have shown human artists to be exceptionally successful at
identifying AI-generated imagery masquerading as human
art~\cite{fakeart-amazon,fakeart-daphne,fakeart-contest}. Instead
of looking for statistical properties of images, human artists look for
inconsistencies in artistic technique, flaws in logic/composition, and other
domain-specific properties that diffusion models do not understand.

Our study evaluates how well skilled artists can use their understanding of
art to detect AI-generated images, by performing separate user studies for 3
separate user populations.
\begin{packed_itemize}
\item{Baseline Participants.} We recruited 180 crowdworkers through the
Prolific online crowdsourcing platform (177 completed and passed attention
checks). Participants were compensated
\$2/10min and this group took on average 8 minutes to complete. This group
included no full-time professional artists and 7 part-time artists.
\item{Professional Artist Volunteers.} We asked for artist volunteers on social media
to participate. Of more than 4000 who responded, 3803
completed the survey and passed all attention checks.
\item{Expert Participants.} We recruited 13 high-profile professional
artists known by members of the research team to have experience
identifying AI imagery. These expert artists are compensated \$25 for
completing the initial survey and detailed feedback, and \$25 more
for participating in the Glaze perturbation user study.
\end{packed_itemize}

\para{Procedure.} The basic user survey included a randomized sample set of
real human artwork, hybrid images, and generative AI images. We ask participants to
classify each image as human-generated, unsure, or AI-generated. We also ask
if they have seen the image displayed before, and answers to previously seen
images are discarded. We ask questions about their artistic expertise, what
styles of art they found easier to classify than others, and factors that
influenced their classification.

We also presented the expert team with a small fixed sample set of AI
generated images that produced the most misclassifications in the other user
studies. In an interactive chat setting, we asked the experts for detailed
feedback on techniques and specific examples applicable to each of these
difficult images.

\subsection{Data Collection}
We curated our own dataset of real human-created artwork, AI-generated
images, and hybrid images. We define real images as original
artwork drawn or created by human artists. AI-generated images are images that are
generated using AI models like Midjourney, Stable Diffusion and
DALL-E 3 from text prompts. {\em Hybrid images} are images that are AI-generated,
retouched, and partially drawn over by humans. One of the
coauthors is a professional artist with over 30 years of
experience. He scanned numerous social media sites and art platforms and 
collected a set of 40 images whose creators admitted they were AI-generated
images altered by human artists later. We describe our data collection
process in detail next.

\section{Constructing the Dataset}
\label{sec:dataset}

We consider images of diverse art styles and sources. We curate a dataset consisting of four different groups of images: artworks handcrafted by human artists (\S\ref{sec:data:human}),  AI-generated images (\S\ref{sec:data:ai}), perturbed versions of human artworks and AI images (\S\ref{sec:data:perturb}), and unusual images created by combining human and AI efforts (\S\ref{sec:data:unusual}).

\secspace
\subsection{Human Artworks}
\label{sec:data:human}

Human artworks are novel creations by artists that capture their personal touch and emotions. They showcase the unique techniques, styles and perspectives of individual artists that only come from years of training and experiences.  With help from the artist community, we collected artworks across 7 major art styles, including anime, cartoon, fantasy, oil/acrylic, photography, sketch, and watercolor.  
We recruited artist volunteers from Cara~\cite{cara}, a major portfolio
platform dedicated to human-created art \annarebuttal{which 
  uses filters to detect AI images and peer-based validation between artists.} 
When recruiting volunteers, we provided artists with a detailed explanation of the study's scope and operations.
We sought their consent to use their artworks in the study and offered them the option to opt out if they were not comfortable with their works being included. 

Overall, we recruited 53 artists and received 280 distinct artworks, mapping to 40 images per style. For each style, we recruited 5 artists specialized for this style and each artist sent us 8 digital images of their own artworks. The  only exception is the watercolor style, where we recruited 7 artists and the number of images sent per artist varies between 4 and 11.

Many artists choose to protect their intellectual property by adding digital signatures or watermarks onto the images of their original artworks. However, many AI-generated images do not have these distinctive marks. Therefore, the presence of signatures or watermarks can potentially influence the perception of human art, introducing unwanted bias and susceptibility to manipulation. To address this issue, we obtained consent from the artists to crop out any signatures or watermarks from their submitted images. In cases where the mark was too adjacent to the art subject, we communicated with the artists to request the original artwork image free of such markings.
Finally, all images were cropped to achieve a square shape, with efforts made to minimize any potential loss of content. This is to maintain consistency between human artworks and AI-generated images, since the latter is of a square shape.

\htedit{\para{Ethics.} Aside from obtaining consent from artists, we take great efforts to minimize exposure of human artworks to external sources.  We provide details in \S\ref{sec:ethics}. }

\secspace
\subsection{AI-Generated Images}
\label{sec:data:ai}
For AI-generated images, we take effort to cover the 7 art styles (listed above) and different AI generators.  We consider the five most popular AI generators: CIVITAI~\cite{civitai},  DALL-E 3, Adobe Firefly, MidjourneyV6, and Stable Diffusion XL (SDXL).  All were the latest release at the time of submission.  For each art style, we prompt each AI generator to produce 10 images,  for a total of 50 images across all five generators.   

\para{Configurating Prompts for Each Art Style.}  We create prompts for AI generators by running BLIP~\cite{li2023blip} on human artworks submitted by artists, generating captions that effectively capture both the artwork's style and content.  BLIP stands out as the state-of-the-art model for image captioning. We apply this method to improve consistency, because artworks of the same style often display large variation in content type and scene. For each art style, we randomly select 10 human artworks, making sure to include at least one piece per contributing artist. The chosen images are input to BLIP to extract the captions. 

Here we encounter an issue where,  for some artworks, BLIP struggles to extract the correct art style or any style at all. For example, for some {\em anime} artworks, BLIP generates captions accurately describing the content but fails to include any style. When prompted by this caption, the AI generators consistently produce images in the {\em photohumanistic} style instead of the intended {\em anime} style,  despite the substantial difference between the two. Similarly, BLIP also fails to extract the watercolor style.  In our study, we address this issue by adjusting the BLIP-generated captions to include the style of the artwork, for which we have ground truth.    Table~\ref{prompts-modifed} in Appendix summarizes the modification made for each art style.

\para{Customizing Prompts per AI Generator.}  We also make customized adjustments on BLIP-generated captions to address unique restrictions and configurations that each AI generator impose on prompts.  Specifically,  Adobe's Firefly and OpenAI's DALL-E 3 models do not respond to prompts that contain certain content.  For instance, Firefly does not generate any image when prompted with ``a fantasy style image of a woman in black holding a knife in the snow,'' but responds properly when the word ``knife'' is replaced with ``sword.'' Similarly,  DALL-E 3 does not react to prompts containing copyrighted materials such as  Marvel character names (e.g., Spider-Man)  and Nintendo game names (e.g., The Legend of Zelda). To address this, we manually substitute such content with more generic terms like `` superhero-themed action figures'' or ``a fantasy-themed action figure on a horse.'' We verify that the modified prompts do produce images that aligned with the intended description.

Another issue is the inconsistent aspect ratio of generated images.  Four out of the five generators consistently produce square images. 
DALL-E 3, on the other hand, generates images with random aspect ratios (e.g. 1024$\times$1792).
DALL-E 3 also tends to self-elaborate on the input prompt, producing extraneous intricacies. To address these artifacts, we include, in each input prompt to DALL-E 3, the additional phrase of ``square image prompt the text to Dall-e exactly, with no modifications.''  Doing so effectively restricts its operation to adhere to the original prompt and return a square image. 

\para{Selecting Art Style from CIVITAI.} Unlike other models, CIVITAI
  hosts instances of SDXL fine-tuned on specific art 
  styles. For each art style, we locate the most frequently downloaded model
  from CIVITAI with that style. For instance, we use ``Anime Art Diffusion
  XL'' to generate {\em anime} style images.

\secspace
\subsection{Perturbed Images}
\label{subsec:dataperturb}
\label{sec:data:perturb}

Users of AI-generated images can intentionally add perturbations to images to
deter their identification as AI images. We consider four representative
types of perturbations and describe each below.  Appendix
(Fig~\ref{fig:perturbation-grid2}) provides visual samples of perturbed images. 

\para{Perturbation \#1: JPEG Compression.} 
\htedit{Existing work has shown that compression artifacts can
  reduce the accuracy of image classifiers~\cite{lam1999effects,
    hamano2023effects, lau2003effects}.  To study its impact on AI image
  detectors, we follow prior work to apply JPEG compression of a quality
  factor 15~\cite{spring2016jpeg} to AI-generated images before querying
  these classifiers.  }

\para{Perturbation \#2: Gaussian Noise.} Similarly, digital noises can be
introduced to disrupt classification-based detectors. For our study, we apply
zero-mean Gaussian noise to each pixel value, with a standard deviation
limited to 0.025, a parameter sweetspot with maximum impact and minimal visual disturbance.

\para{Perturbation \#3: CLIP-based Adversarial Perturbation.} \htedit{A more advanced (and costly) approach is to apply adversarial perturbations on AI images. Adversarial perturbations~\cite{goodfellow2014explaining,madry2017towards} are carefully crafted pixel-level perturbations that can
confuse ML classifiers. Automated AI image detectors are known to rely on the public CLIP model~\cite{cozzolino2023raising,sha2023defake} for detection, and thus, we leverage the CLIP model to craft 
our adversarial perturbations to maximize their transferability to AI detectors~\cite{demontis2019adversarial}. Specifically, we compute LPIPS-based adversarial perturbation~\cite{ghazanfari2023r} on each AI-generated image. We ensure that the perturbation is sufficient to confuse the CLIP model
(i.e., LPIPS budget = 0.03). }

\para{Perturbation \#4: Glaze.} \htedit{Glaze~\cite{shan2023glaze} is a tool for protecting human artists from unauthorized style mimicry. It introduces imperceivable perturbations on each artwork, which transforms the image's art style to a completely different one in the feature space.  The widespread use of Glaze by artists has sparked extensive online discussions focused on instances where the use of Glaze on human art results in detection as AI images, while applying Glaze on AI-generated images can evade detection.  To understand its impact, we use the public WebGlaze~\cite{webglaze} tool to perturb both human art and AI images. We choose both the default medium intensity and also the high intensity, as artists often employ strongest protection to safeguard their online images. }

\secspace
\subsection{Unusual  Images}
\label{sec:unusualdata}
\label{sec:data:unusual}
\para{Hybrid Images.}   
\htedit{Users can create ``hybrid'' images by painting over AI-generated images. When posting them online, many include in the caption the generative models used.  One of the coauthors, a professional artist over 30 years of experience, collected 40 hybrid images to include in our dataset and verified their sources.} The images cover a variety of styles and subjects, including anime, cartoon, industrial design, and photography. 

\para{Human Artworks with Upscaling.} Some artists use tools like image upscalers to enhance the quality of photography images, e.g., reducing blur or noise introduced during image capturing.  We have 70 images in this group, upscaled using the baseline function of MagnificAI~\cite{magnificai}, a web upscaling tool endorsed by artists.

\section{Accuracy of Automated Detectors}
\label{sec:MLresults}

Using the dataset outlined in \S\ref{sec:dataset}, we 
examine the efficacy of automated detectors in detecting AI-generated
images. We first report the results on unperturbed imagery and the
impact of AI generator choice.  We then consider advanced scenarios
where the detectors face different types of perturbed images,  benign or malicious.

\secspace
\subsection{Experiment Setup}
\label{sec:MLsetup}
As discussed in \S\ref{sec:setup}, we consider five
classification-based detectors,
including three commercial detectors (Hive, Optic, Illuminarty),  and
two detectors built by 
academic researchers (DIRE, DE-FAKE).  Our experiments
use both original, unperturbed imagery (280 human artworks and 350 AI-generated
images) and their perturbed versions.  We delay the study of unusual 
images to \S\ref{subsec:unusualresult} due to their decision
complexity.

\para{Detector Decisions.} We study the 
ability of automated detectors to identify a human artwork as human and
an AI-generated image as AI,  mapping to a binary decision. However, today's automated detectors
all output a probabilistic score indicating the likelihood or confidence
of the input being AI-generated. A score of 100\% implies absolute
certainty that the input is AI-generated, while 0\% indicates it is
definitely human art. To convert this score into a binary
decision, we need to establish a boundary that distinguishes between the two classes. Relying on a single threshold, such as $x$\% (e.g., 50\%), is obviously too fragile for this purpose.

Instead, we leverage the widely used 5-point
Likert scale in user studies~\cite{jeong2016level}, designed to obtain quantitative
measures of perception/decision. Specifically, scores ranging from 0-20\% are
associated with category human artwork (very confident), 20-40\% with
human artwork (somewhat confident), 40-60\% with ``not sure,'' 60-80\%
with AI-generated (somewhat confident), and 80-100\% with AI-generated
(very confident). It is important to note that our user studies
maintain consistency by adopting the same rating scale. Next, to produce binary decisions,  we designate any score below 40\% as a
decision of human art and any score above 60\% as a decision of
AI-generated.  Those inputs yielding scores between 40-60\%,
indicating uncertainty (``not sure''), are excluded from the 
experiment.  This exclusion is based on two practical
considerations. First,  there is no equitable method for comparing
a ``not sure'' decision against the definitive ground
truth~\cite{chyung2017evidence}. Second, the occurrence of ``not sure'' is  minimal across
the machine detectors, e.g., 0.54\% for Hive and 4.29\% for Optic,
thus their removal has a negligible impact on overall 
performance.

\para{Evaluation Metrics.} We report detector performance using four
metrics.  For easy notation, let [H$\rightarrow$H] represent the
\# of human artworks detected as human-made, and
[AI$\rightarrow$AI] represent the \# of AI-generated images 
detected as AI-generated. Let [H] and [AI] represent the total
\# of human artworks and AI-generated images included in this test, respectively. 

\begin{packed_itemize}\vspace{-0.05in}

\item {\bf Overall accuracy (ACC)} measures the accuracy of the detector
  regardless of the data origin. ACC= ([H$\rightarrow$H]+[AI$\rightarrow$AI])/([H]+[AI]). 

\item {\bf False positive rate (FPR)} represents the ratio of human 
  artworks misdetected as AI-generated.  FPR=1- [H$\rightarrow$H]/[H]. 

\item {\bf False negative rate (FNR)} measures the ratio of AI-generated
  images misdetected as human artworks. 
  FNR=1 - [AI$\rightarrow$AI]/[AI]. 
  
\item {\bf AI Detection success rate (ADSR)} captures the
  detector accuracy on AI-generated images. 
  ADSR= [AI$\rightarrow$AI]/[AI].   We use this metric to 
   examine how generation-related factors 
  would affect the detection outcome of AI-generated images. 
  \vspace{-0.05in}
\end{packed_itemize}

\secspace
\subsection{Results of Unperturbed Imagery}
We start from unperturbed human artworks (280 images) and
AI-generated images (350 images). Table~\ref{tab:core-detector-table} 
summarizes the detection performance in
terms of ACC, FPR and FNR.

\para{Hive.} Hive is the clear winner among all five detectors, with a
98.03\%  accuracy, 0\% FPR (i.e., it never misclassifies human
artworks), and 3.17\% FNR (i.e., it rarely misclassifies AI-generated
images).

\para{Optic and Illuminarty.} Both perform worse than Hive, except 
that Optic has a lower FNR (1.15\%) and thus is more effective at
flagging AI-generated images. However, this comes at the expense of a
high 
24.47\% FPR, where human art is misclassified as AI-generated.
Illuminarty demonstrates even harsher treatment towards human art,
with a very 
high FPR of 67.4\%.

\para{DE-FAKE and DIRE.}  We experiment with all 6 versions of DIRE, representing
its 6 checkpoints published online. The detailed performance is
listed in Table~\ref{tab:dire-checkpt} in Appendix.  The overall
accuracy is consistently low for all 6 models ($<55.5$\%).  The top-2 models' performances are
listed in Table~\ref{tab:core-detector-table} as DIRE (a) and DIRE
(b), one with nearly 100\% FPR, and another with 66\% FNR. DE-FAKE shows a similar pattern, with a low 50\% accuracy and high
FPR and FNR values.   Given their poor performance, we do not conduct
any additional experiments with both detectors.

\begin{table}[t]
    \centering
    \resizebox{0.36\textwidth}{!}{
    \centering
\begin{tabular}{c|ccc}
        \multicolumn{4}{c}{Tested on Human Artworks + AI-generated Images } \\ \hline
Detector & ACC (\%) $\uparrow$ & FPR (\%) $\downarrow$ & FNR (\%) $\downarrow$ \\ \hline
{\bf Hive}        & {\bf 98.03}    & {\bf 0.00}      & {\bf 3.17}     \\
Optic       & 90.67    & 24.47     & 1.15  \\
  Illuminarty & 72.65    & 67.40     & 4.69 \\
  DE-FAKE &50.32   & 41.79 & 56.00 \\
  DIRE (a)       &   55.40       &   99.29  & 0.86 \\
  DIRE (b)       & 51.59         &25.36  & 66.86 \\ \hline
   {\bf Ensemble} & {\bf 98.75}                        & {\bf 0.48 }
                                                       &  {\bf 1.71}                         \\
  \hline
\end{tabular}}
\caption{Performance of automated detectors tested on
  unperturbed human artworks and AI-generated images. The Ensemble
  detector (Hive+Optic+Illuminarty) takes scores from Hive, Optic, and Illuminarty, using the highest confidence
  value as the determining score.}
\label{tab:core-detector-table}
\vspace{-0.2in}
\end{table}

\para{Variation across Automated Detectors.} The large performance
variations across detectors could be attributed to the
diversity and coverage of their training data.  According to its website, Hive utilizes a rich collection of generative AI
datasets and can identify the 
generative model used for the current input from a pool of nine models
\annarebuttal{\footnote{As of April when conducting additional tests on Hive, it now detects 27 models.}}. 
Similarly, Optic can pinpoint an input image
to one of the 4 generative models. In contrast, Illuminarty's
training data coverage is limited, particularly since it does not
support image files exceeding 3MB. In our study, we have to
downsample images generated by Firefly from 2048x2048 to 1024x1024 to
meet this restriction. \annarebuttal{Illuminarty and Optic are smaller companies with less training data 
\footnote{Illuminarty's creator informed us the model has not been updated for 6 months in Novemeber 2023.}.}

DE-FAKE and DIRE's training data are constrained and
lack representation from
art. 
DIRE models are trained on interior design images
(LSUN-Bedroom), human faces (CelebA-HQ) and ImageNet. The detection accuracy on art images is 
around 50\%.  For DE-FAKE,  we follow the methodology described
  by~\cite{sha2023defake} to train the classifier using images generated by SDXL and
MSCOCO captions.  When tested on
SDXL images produced from MSCOCO prompts,  the classifier reproduces
a high ACC of 92.44\%  similar to~\cite{sha2023defake}. 
However, when tested on images generated using artwork
prompts, the accuracy drops to 50.3\% for SDXL 
images and 46.43\% for those produced by other generators.
\annarebuttal{We attribute performance discrepancy to the issue of transferability. 
Training images generated from MSCOCO prompts do not follow the art dataset distribution, 
so these open-source detectors did not transfer well to the art dataset.}

  \para{Combining Detectors (Hive+Optic+Illuminarty).}  \jpedit{We also study an Ensemble detector that leverages
  decisions from Hive, Optic, and Illuminarty. In case of a disagreement between the detectors, it opts for the
  decision marked with highest confidence. The confidence is calculated relative to the classification, by computing $|$ detector score - 0.5 $|$. The result in 
  Table~\ref{tab:core-detector-table} shows that the improvement over
  Hive is minor:  0.6\% increase in ACC while FNR reduces from 3.17\%
  to 1.71\% and FPR increases from 0\% to 0.48\%. 

}

\begin{table}[t]
    \centering
    \resizebox{0.42\textwidth}{!}{
    \centering
    \begin{tabular}{c|ccccc}
        \multicolumn{6}{c}{ADSR (\%) $\uparrow$}\\ \hline
 & CIVITAI & DALL-E 3 & Firefly & MJv6  & SDXL   \\ \hline
Hive     & 100.00  & 98.57  & 91.04   & 94.29 & 100.00 \\
Optic       & 100.00  & 97.14  & 97.06   & 100.00& 100.00 \\
Illuminarty  & 94.03   & 100.00  & 92.42   & 91.67 & 98.41 \\\hline
\end{tabular}}
\caption{The impact of AI-generator choice on detector performance,
  represented by ADSR, the \% of AI-generated images correctly
  detected as AI-generated. }
\label{tab:ADSR-generator}
\vspace{-0.25in}
\end{table}

\secspace
\subsection{Impact of AI Generator Choice}
\begin{figure}[t]
    \centering
    \includegraphics[width=0.45\textwidth]{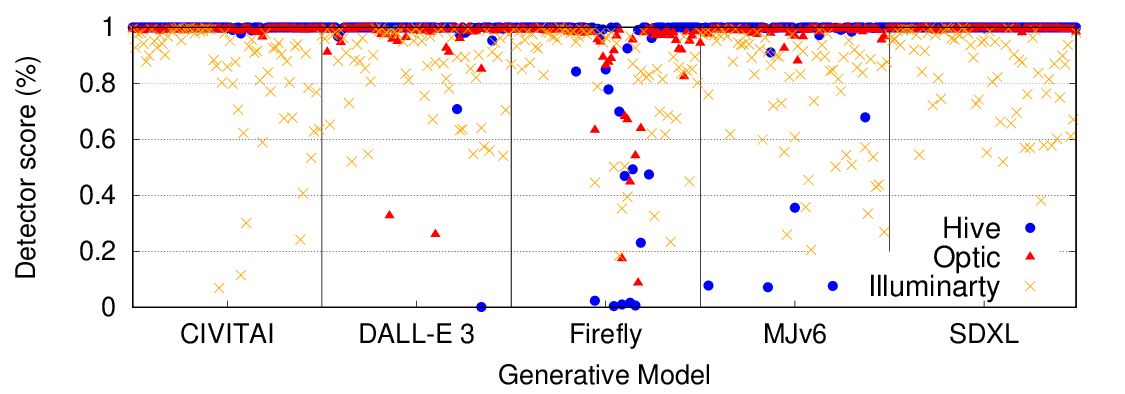}
        \vspace{-0.1in}
    \caption{The confidence score produced by automated detectors on
   images generated by 5 generators.  Detecting images
      generated by Firefly is the hardest.}
    \label{fig:generative_model_accuracy}
    \vspace{-0.1in}
\end{figure}

We also investigate the factors that may affect detection
performance. While one might anticipate that the art style 
could have an impact, we do not observe any notable
effect, at least for the seven major styles considered by our
study. Instead, we observe a considerable impact by the choice of
AI-generator.  This can be seen from both
Table~\ref{tab:ADSR-generator}  and
Figure~\ref{fig:generative_model_accuracy}. 
Table~\ref{tab:ADSR-generator} lists ADSR,  the \%
of AI-generated images correctly detected as AI, while
Figure~\ref{fig:generative_model_accuracy} displays the
raw confidence score produced by the detectors.

Across the five AI generators, images by CIVITAI and SDXL are the ``easiest'' to
detect by Hive and Optic, i.e., 100\% ADSR and not a single ``not
sure'' decision. \anna{Since CIVITAI models are fine-tuned versions of
SDXL, this suggests fine-tuning has minimal impact on AI image detection.}  

On the other hand, Firefly images are the least detectable -- Hive
marks  6 out of 70 Firefly images as human art  and 3
as ``not sure'' while Optic marks 2 as human art and 2 as ``not
sure'' (Fig.~\ref{fig:generative_model_accuracy}).

\anna{We hypothesize this is due to lack of training data. Firefly is a relatively new
  model, so Hive and Optic are likely to have much less training
  data relative to other models}

\secspace
\subsection{Impact of Perturbations}
\label{sub:perturbresult}

Our goal is to understand whether adding perturbations to images,
whether benign or malicious, could
change detection outcomes. This triggers two questions below.

\begin{figure*}[t]
  \centering
  \includegraphics[width=1.4\textwidth]{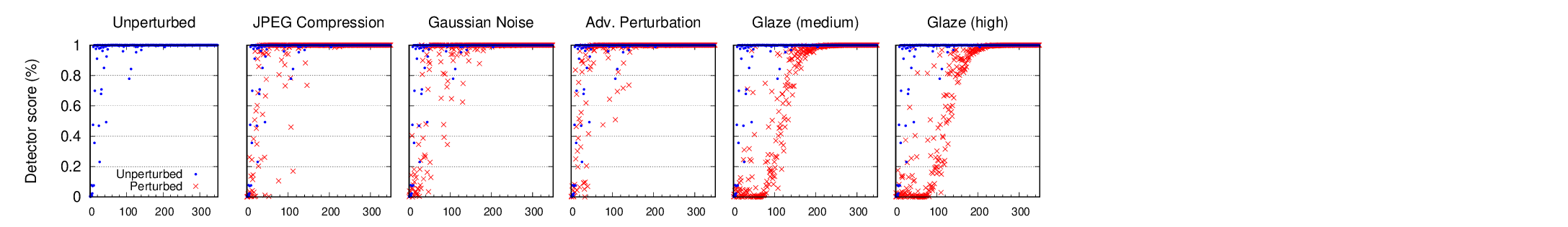}
  \vspace{-0.2in}
  \caption{Impact of five different perturbations on the Hive
    confidence score,  for 350 AI-generated images. In each figure, the images are indexed by the
    increasing Hive score of unperturbed versions.} \vspace{-0.1in}
  \label{fig:perturbation_hive_multi}
\end{figure*}

  \begin{table}[t]
  \centering
  \hspace{-0.2in}
    \resizebox{0.5\textwidth}{!}{
    \centering
    \begin{tabular}{c|ccccc|c}
          \multicolumn{7}{c}{ADSR (\%) $\uparrow$}\\ \hline
 & \begin{tabular}[c]{@{}l@{}}JPEG\\ comp.\end{tabular}  & \begin{tabular}[c]{@{}l@{}}Gaus.\\ noise\end{tabular} & Adver.  &  \begin{tabular}[c]{@{}l@{}}Glaze\\(med.) \end{tabular} & \begin{tabular}[c]{@{}l@{}}Glaze\\(high) \end{tabular}  & Unperturbed \\ \hline
Hive             &  91.88  & 88.73     & 93.00    &   69.73              & 67.56             & 96.83 \\
Optic          &   97.98   &  52.63   & 80.42    &   62.62               & 58.00          & 98.85 \\
Illuminarty     &  93.19  & 61.43    & 94.34    &   89.80                &89.22            & 95.31 \\ \hline 
Ensemble & 94.29 & 88.86 & 94.29 &  73.71 & 71.43 & 98.29 \\\hline
    \end{tabular}}
\caption{The impact of perturbations on AI-generated
  images,
  represented by ADSR, the \% of AI-generated images correctly
  detected as AI-generated.}
\label{tab:perturbations-table}
    \vspace{-0.2in}
\end{table}

\para{Question 1: How do perturbations affect the detection of AI images?}
\jpedit{
We explore four perturbations: JPEG compression, Gaussian noise, adversarial perturbations on CLIP, and Glaze at medium and high intensity.
The details of each perturbation are discussed at length in Section \S\ref{sec:data:perturb}.

Table~\ref{tab:perturbations-table} reports ADSR, the success rate of detecting AI-generated images as AI when they include one of the four perturbations. 
As a reference, we include the ADSR for unperturbed images. 
We discuss each of the perturbations individually below, analyzing their impact on each detector.

Figure~\ref{fig:perturbation_hive_multi} plots a detailed view of the impacts of perturbations on the scores assigned by Hive to each image. 
While we examined these plots for each detector, we observed relatively similar trends across all, except that Optic and Illuminarty exhibited higher levels of noise.
Hive's plots are easier to interpret, both for the perturbed and unperturbed data, thus we only present them for clarity. Additionally, we identify a set of images that demonstrate `robustness' to any perturbation against Hive and provide a more in-depth discussion of these images in Table~\ref{tab:hive-robust-table} in Appendix.

\para{JPEG Compression.} JPEG compression shows minimal impact on performance across all detectors, as they all remain above 91\%. 
The lossy compression artifacts don't hinder the detectors' ability to detect AI images.

\para{Gaussian Noise.}  Gaussian noise has little impact on Hive, it does
drop both Optic and Illuminarty in ADSR.  Optic's ADSR decreases to 52.63\%,
and Illuminarty's ADSR decreases to 61.43\%, both of which are 
near random guessing.  On the other hand, Hive's performance remains
relatively high at 88.73\% ADSR. \anna{These classifiers may have already adapted to the presence of mild noise in training images and learned to suppress the effects of noise.}

\para{Adversarial Perturbation.} Of all the perturbations, the adversarial perturbations on CLIP have the least impact on performance for Hive and Illuminarty. 
Optic's ADSR drops to 80.42\%, but Hive and Illuminarty both remain above 93\%.
This may indicate relatively low transferability of targeted attacks in the CLIP space in black-box settings. 

\para{Glaze.} Across detectors, Glaze at both intensities consistently has a significant impact on ADSR.
Additionally Glaze affects each detector similarly between medium and high intensity, with the increase in intensity only 
resulting in a drop in ADSR ranging from 0.5\% to 4\%. Glaze has the least effect on Illuminarty, 
dropping the ADSR from 95.31\% to 89.22\% for high-intensity Glaze. 
In comparison, Glaze reduces the ADSR for both 
Hive and Optic to below 70\% while the Ensemble detector is only able to achieve 71.43\%. 
We explore the effects of Glaze further below. 
}

\para{Question 2: Does Glaze Affect Accuracy Similarly on Human Artwork vs AI-generated Images?}
To investigate if the detection of human artwork is impacted similarly to AI-generated images when Glazed, 
we applied high-intensity Glaze to both. 
The detection outcomes, including ACC, FPR and FNR are presented in Table~\ref{tab:glaze-table} comparing results with and without Glaze.
Our findings reveal a large reduction in ACC with the use of Glaze, as expected. 
However, the FNR increases across all detectors, while the FPR remains relatively consistent. 
On Hive, ACC is decreased around 20\% and yet FPR increases only 3.23\% while FNR increases almost 30\%. 
Our Ensemble detector is able to achieve the highest accuracy on Glazed images, but maintains a FNR of 28.57\%. 
This suggests that glazing human artwork typically does not affect classification success, 
but glazing AI-generated images often leads to misclassification as human artwork.  
We attribute this to the scarcity of Glazed images online and thus the lack of Glazed images in the detector's training datasets. 
Additionally Glaze was created to protect human artwork and has since been adopted by many artists, 
therefore the distribution of Glazed human art vs. Glazed AI-generated images online is likely skewed. 
Once again this demonstrates the impact of insufficient training data on image-based classifiers.

\begin{table}[t]
    \centering
    \resizebox{0.4\textwidth}{!}{
    \centering
\begin{tabular}{c|ccc}
&  ACC (\%) $\uparrow$ & FPR (\%) $\downarrow$ & FNR (\%) $\downarrow$  \\ \hline
Hive        & 80.81 / 98.03     & 3.23 / 0     & 32.44 / 3.17     \\
Optic       & 61.92 / 90.67     & 33.59 / 24.47   & 42.00 / 1.15    \\
Illuminarty & 68.66 / 72.65     & 56.91 / 67.40    & 10.78 / 4.69  \\ \hline 
Ensemble & 82.70 / 98.75     & 3.21 / 0.48    & 28.57 / 1.71  \\ \hline 
\end{tabular}}
\caption{Detection performance on  human art and
  AI-generated images with and
without Glaze (at high intensity), shown as: Glazed / Unperturbed.}
\label{tab:glaze-table}
\vspace{-0.2in}
\end{table}

\secspace
\subsection{Summary of Findings}
Our study leads to three key findings.
\vspace{-0.05in}
\begin{packed_itemize}
\item Commercial detectors perform surprisingly well, and Hive performs the best.  
It is highly accurate (98.03\% accuracy) and never misclassifies human artwork in our test. 
The other two detectors (Optic and Illuminarty) tend to misclassify human art as AI.

\item Commercial detectors are heavily affected by feature space perturbations (i.e., Glaze) added to AI-generated images. 
On the other hand, human artworks with Glaze are mostly unaffected, as they are still largely detected as human.

\item Poor performance on Firefly and Glaze indicates the results are
  correlated with training data.  Performance of supervised
  classification depends heavily on the availability of training
  data. Detectors are more vulnerable to newer models with less available
  training data (Firefly) and adversarial inputs they have not seen before.

\end{packed_itemize}

\begin{figure*}[t!]
  \centering
  \begin{minipage}{0.35\textwidth}
    \centering
    \includegraphics[width=1\columnwidth]{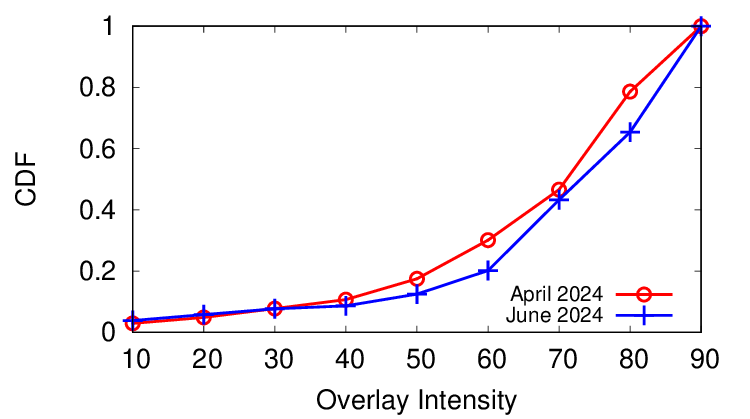}
  \caption{CDF of overlay intensity required to change Hive's decision over the period of 3 months.}
  \label{fig:cdf_month_comparison}
  \end{minipage}
  \hfill
  \begin{minipage}{0.35\textwidth}
    \centering
    \includegraphics[width=1\columnwidth]{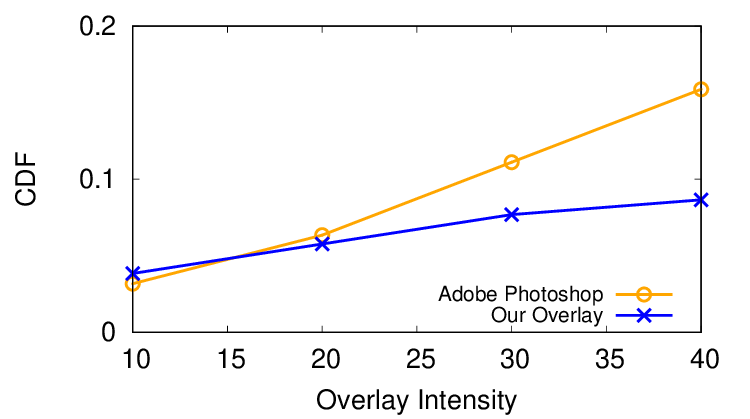}
  \caption{CDF of overlay intensity required to change Hive's decision using different overlay methods in June 2024.}
  \label{fig:cdf_ours}
  \end{minipage}
  \hfill
  \begin{minipage}{0.19\textwidth}
    \centering
    \includegraphics[width=0.88\textwidth]{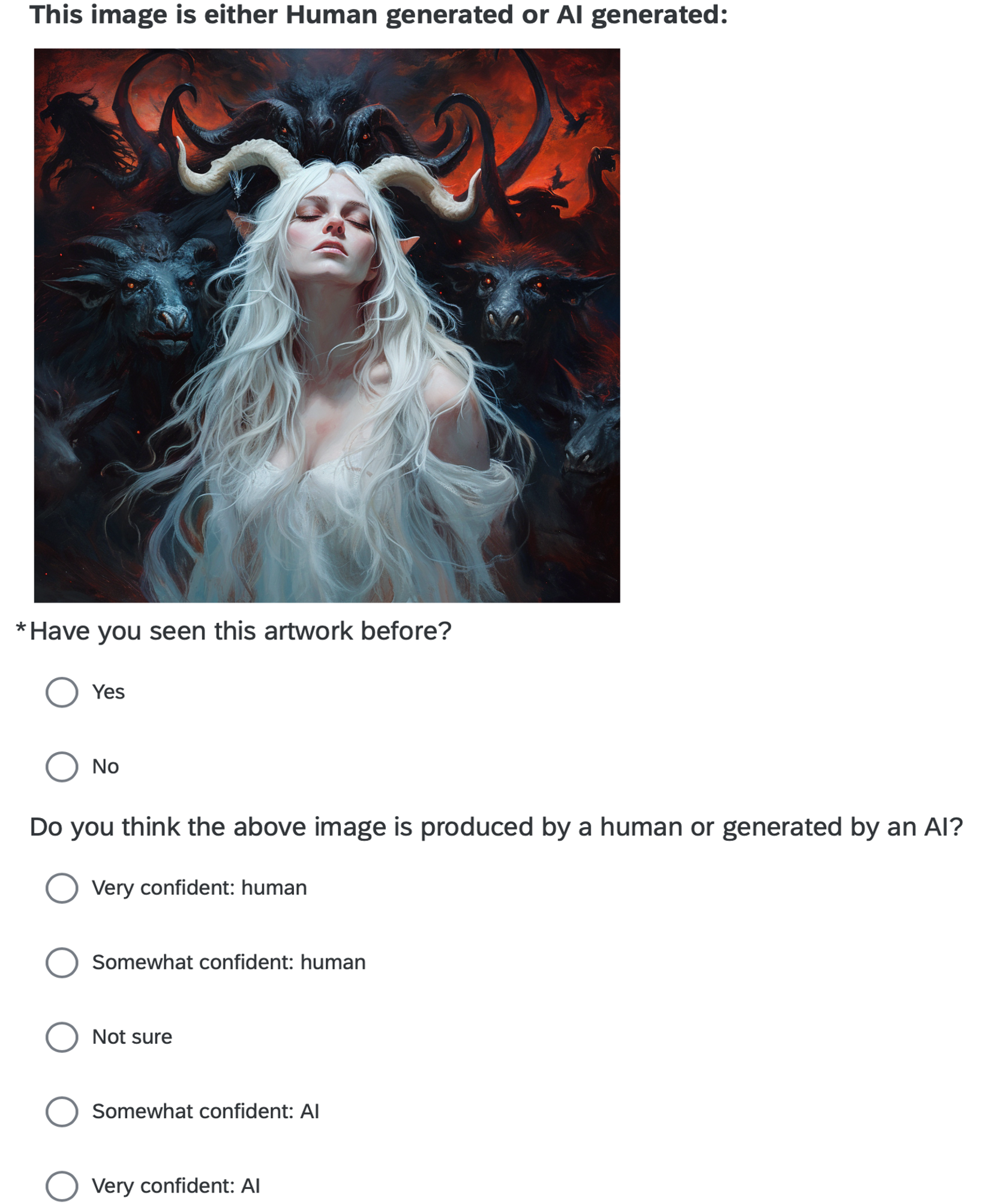}
    \caption{User study interface.}
    \label{fig:userstudyinterface}
  \end{minipage}
\end{figure*}

\annarebuttal{
\secspace
\subsection{Followup Robustness Tests on Hive}
After the paper was accepted, we discovered a Reddit post that claimed
overlaying a real image of a white wall onto an AI-generated image could
bypass Hive's detection~\cite{reddithive}. We investigate this as a new
adversarial approach to bypassing detection, and conduct a large-scale test
on 105 AI-generated images, randomly sampled from our dataset to include 3
images from each of the 7 styles and 5 AI-generators.  We implement a python
script to overlay each image with a white wall image from Adobe
Shutterstock, at intensity levels varying from 10\% to 90\% in increments of
10.
Hive confirms the Adobe's white wall image as not AI-generated with
100\% confidence.

Figure~\ref{fig:cdf_month_comparison} illustrates the CDF graph of the
distribution of overlay intensities required to flip Hive's decision from
AI-generated to not AI-generated. The red curve from April 2024 shows there
is a 17.5\% chance that Hive's decision will change at or below intensity
levels of 50\%.  Yet, there are corner cases where images (2.91\% in
April and 3.85\% in June) were able to fool Hive with the white wall 
overlaid at only 10\% intensity. Most images require around overlay intensity
of 60-80\% to change Hive's classification.

We also note that Hive's performance improved over time.
Figure~\ref{fig:cdf_month_comparison} displays Hive results collected in
April 2024 and June 2024. The shift in the CDF shows that Hive's accuracy and
robustness increased in the 2 month period, possibly due to adversarial
training on inputs or updates to its supervised classification algorithm.

We conduct another smaller-scale test to see the effect of different overlay
methods on Hive's output.  We apply and compare Adobe Photoshop's ``multiply
blend'' overlay option (the method used in the Reddit post) against our
scripted blending algorithm, varying intensities from 10\% to 40\% in
increments of 10 across 63 AI-generated images\footnote{Dataset is narrowed
  down to 3 images from MJv6, Firefly, and SDXL across 7 styles.}.
Figure~\ref{fig:cdf_ours} shows Adobe's Photoshop method has a significantly
stronger effect against Hive at higher intensities. This gap is somewhat
unexpected, and just confirms that even small variations in adversarial
attacks can have large unpredictable impact on Hive's detection robustness.}

\secspace
\section{User Studies on Human Detection}
\label{sec:userbase}
Next, we measure the ability of human users and artists in identifying
AI-generated images.  As discussed in \S\ref{subsec:humansetup},  we
perform user studies with 3 separate user populations, including 
crowdworkers, professional artists and expert artists. 

To ensure uniformity, each group is provided with the same user study, but
the expert group received extended followup questions asking for detailed
examples (Section~\ref{subsec:experts}).  Our study is approved
by our local Institutional Review Board (IRB). We omit the IRB number for
anonymity.

\secspace
\subsection{Study Setup} 

\para{Participants.} As discussed in \S\ref{subsec:humansetup}, we recruited
three participant groups, and a total of {\bf 3993} participants completed
our study and passed all attention checks.  These include 177 baseline
participants recruited from Prolific, referred to as general users,
3803 professional artist volunteers, and 13 high profile experts.

\para{Task.} Our study takes the form of a user survey.  We added
attention-check questions in both the middle and concluding sections of the
survey to filter out low attention participants.

After a brief introduction on generative models and the current issue of
distinguishing between human artworks and AI-generated images, we present 
the participant with a sequence of 20 images, shown one at a time, and ask
them to decide whether each image is human-made or AI-generated.
For each image, we ask two questions. The first question is ``Have you seen this
image before?''  If yes, we disregard the response for this image, since the
participant has likely seen this image and is possibly aware of the image's true source.  The 
second question asks them to rate the current image with one of the five
choices: ``human art (very confident),'' ``human art (somewhat confident),''
``not sure,'' ``AI-generated (somewhat confident),'' ``AI-generated (very
confident).''  For each participant, we randomly sample 20 images from a
collection of 670 images, consisting of 210 human artworks, 70 human artworks
after upscaling, 350 AI-generated images, and 40 hybrid images. \annarebuttal{Every image is seen by five participants.}

Next, we ask the participant, for each of the 7 art styles, whether they are
confident at distinguishing human art from AI-generated images; \annarebuttal{the user study interface is presented in Figure~\ref{fig:userstudyinterface}.}
If confident, we ask them to describe the properties of the art that contributed to their decisions.
Here we present seven options: ``content,'' ``complexity,''
``technical skill,'' ``perspective,'' ``lighting,'' ``consistency,'' and an
additional text response section for additional details. 
Finally, we ask them if they self-identify as full-time or part time artists. 

\para{Performance Metrics.} Same as the evaluation of automated
detectors in \S\ref{sec:MLsetup},  we convert each  5-point Likert
scale rating into a binary decision. That is, both ``human art (very
confident)'' and ``human art (somewhat confident)'' map to a decision
of human art, and ``AI-generated (somewhat confident)'' and
``AI-generated (very confident)'' map to AI-generated.  Not decision
is produced for the ``not sure'' responses, and we ignore them when
computing ACC, FPR, FNR and ADSR (same as \S\ref{sec:MLsetup}).

\secspace
\subsection{Detection Accuracy for General Users} 
Table~\ref{tab:human-ACC} shows detection performance of general users, which
is only slightly better than random coin-flip. This shows that general,
non-artist users are unable to tell the difference between human art and
AI-generated images.

Table~\ref{tab:style-table} examines the impact of art style on
detection accuracy (ACC). For general users, accuracy is slightly
higher on cartoon, fantasy and photography styles. These three
represent a collection of ``digital'' artworks more frequently
accessible to general users, compared to ``physical'' art styles
like oil/acrylic, watercolor, and sketch, and other ``digital''
artworks like anime.  We attribute the slightly higher accuracy 
to this increased familiarity. 
Finally, Table~\ref{tab:human-ADSR} reports detection success rate
on AI-generated images, which is around 60\% and varies
slightly across the five AI generators.  Images generated by Firefly
and MidjourneyV6 are harder for non-artist users to recognize.

\secspace
\subsection{Detection Accuracy for Professional Artists}
\label{subsec:profresult}

Compared to general users, professional artists take more time to inspect images and are more
effective at distinguishing between human art and AI
images. They produce a detection accuracy of 75.32\%
(Table~\ref{tab:human-ACC}). Their 23.53\% FPR and
25.37\% FNR indicate that  their detection performance is not skewed toward
either human art or AI-generated images.

\para{Decision Factors.}  To understand why professional artists are
better at evaluating art images,  we study the key factors that
influenced their decisions.  The first is the image's art style.
Table~\ref{tab:style-table} shows that the image's artistic style has
a clear impact on the performance of professional artists.   Among the
7 styles, the top-3 ``easier-to-detect'' styles are anime, cartoon and
fantasy, and the bottom one is oil/acrylic for which the detection
accuracy drops nearly 20\%.   This aligns with the participants'
feedback on the styles that they feel most confident on detection,
where 82.42\% selected cartoon, followed by fantasy and anime.

For these top-3 styles, the artists select ``consistency'' as the most
dominant decision factor. Specifically, in the text response section on anime
images, the most frequently entered words are hands, hair, details, eyes and
lines. Similarly, in the case of fantasy images, artists observe specific
details such as asymmetric armor or symbols that should be symmetric.  This
shows that professional artists can apply their knowledge and experiences of
art creation to identify inconsistencies in AI-generated images, with
particular focus on fine-grained details.

\para{Impact of AI generator.} Table~\ref{tab:human-ADSR} lists the detection
success rate of AI-generated images broken down by source generator model.
For professional artists, accuracy clearly varies across generator models.
The top-2 ``easiest-to-detect'' are CivitAI and SDXL, with detection
success rate above 83\%.  Interesting to note that these two are also the most
easy-to-identify generators for automatic detectors. Next, detection
accuracy reduces to below 70\% for images produced by MidjourneyV6 and DALL-E
3, suggesting that MidjourneyV6 and DALL-E 3 produce better copies of human
art styles, making it harder for artists to spot inconsistencies.

\begin{table}[t]
    \centering
    \resizebox{0.4\textwidth}{!}{
    \centering
\begin{tabular}{c|ccc}
& ACC (\%) $\uparrow$ & FPR (\%) $\downarrow$ & FNR (\%) $\downarrow$ \\ \hline
  General user & 59.23  & 40.81     &  40.75     \\
	Professional artist  &  75.32   & 23.53      & 25.37 \\
  Expert artist & 83.00    & 20.78     & 14.63 \\ 
  \hline
\end{tabular}}
\caption{Performance of human detection on both human art and AI-generated images.} 
\label{tab:human-ACC}
\vspace{-0.2in}
\end{table}

\begin{figure*}[t]
  \centering 
  \includegraphics[width=2\columnwidth]{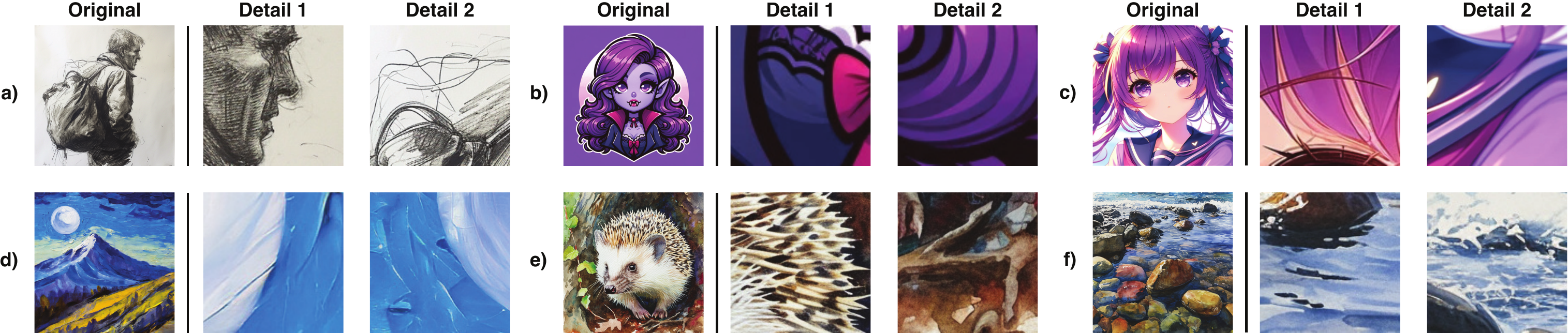}
      \vspace{-0.1in}
      \caption{Six hard-to-detect AI-generated images, and their
        artistic errors/inconsistencies discovered by our expert
        artists.  In each 3-image group, the left one is the full
        image, and the right two are zoomed-in view of discovered
        artifacts. } 
  \label{fig:hard-ai}
  \vspace{-0.1in}
\end{figure*}

\begin{table}[t]
  \centering
  \resizebox{0.49\textwidth}{!}{
  \centering
\begin{tabular}{c|ccccccc}
       & Anime & Cartoon & Fantasy & Photo. & Oil/Acrylic & Sketch & Watercolor \\ \hline
Gen. & 57.43 & 63.58   & 62.80   & 63.23       & 56.81       & 54.99  & 56.16      \\
Pro. & 79.16 & 82.42   & 81.72   & 76.05       & 66.01       & 70.27  & 70.51     \\ \hline
\end{tabular}}
\caption{Impact of art style on detection accuracy (ACC) for
  general users (Gen.) and professional artists (Pro.). We exclude the result of expert artists due to insufficient coverage.}
\label{tab:style-table}\vspace{-0.25in}
\end{table}

\begin{table}[t]
  \centering
  \resizebox{0.46\textwidth}{!}{
  \centering
  \begin{tabular}{c|ccccc}
     \multicolumn{6}{c}{ADSR (\%) $\uparrow$}\\ \hline
       & CIVITAI & DALL-E 3 & Firefly & MJv6  & SDXL   \\ \hline
General user  & 66.77  & 60.63  & 51.18   & 50.00 & 67.56 \\
Professional artist   & 83.56  & 67.56  & 75.40   & 61.53& 84.43 \\
Expert  artist  & 90.32   & 86.96  & 65.22   & 86.96 & 95.65 \\\hline
\end{tabular}}
\caption{ADSR of human detection on AI-generated images by different AI generators. ADSR is the success rate of detection AI-generated images as AI-generated. 
}
\label{tab:human-ADSR}
\vspace{-0.2in}
\end{table}

\secspace
\subsection{Detection Accuracy for Expert Artists}
\label{subsec:experts}

Our 13 expert artists show greater proficiency in the detection
task, raising overall detection accuracy (ACC) to 83\% (see 
last row in  Table~\ref{tab:human-ACC}).  They also produce slightly
imbalanced FPR (20.78\%)  and FNR (14.63\%), indicating that they
are better at spotting AI-generated images than human art. When we gave
feedback to the experts on their results, many were frustrated that they
committed errors by marking human art as AI (false positive). In retrospect,
they explained that they identified detailed mistakes and inconsistencies
which were likely due to inexperience and human error by the human
artists. For example, in a painting of a bedroom bathed in moonlight, the
shadow of a window pane had slightly offset position of the latch compared to
the window itself. This was seen by expert artists as an inconsistency, but
later attributed to lack of attention to detail by a human artist. 
\annarebuttal{This attests for the drop in ADSR for Firefly images. Since expert artists look at fine-grained detail in AI-generated images,
they are overfitted to spot irregularities from popular generators and have yet been accustomed to the newer style of images.}

\para{Decision factors.} Our expert artists are generally very confident at
judging images from more than 2 art styles. The most frequently selected one
is fantasy art. \htedit{The primary decision factors are intrinsic artistic
  details, which often go beyond the ``(in)consistency'' element used by
  professional artists in their detection efforts (as discussed in
  \S\ref{subsec:profresult}).} Specifically, our expert artists point out
that AI-generated images generally ``look too clean, rendered, and detailed''
and ``have no variety in composition, edges, distribution of detail,'' and
had ``design elements are nonsensical or blend into each other in telltale ways,''  
while human-made fantasy art ``contains components that are novel such as
armor or jewelry.''

\para{Focus Study on ``Hard-to-Detect'' AI Images.}  For a comprehensive view
of how experts identify AI-generated images, we presented our expert group
with a fixed set of the six AI-generated images that produced the most false
negative errors by the professional (non-expert) artist group.  They cover 5
styles: sketch, oil/acrylic, cartoon, watercolor, and anime.  For each of
these difficult images, we asked the experts for detailed feedback on what
exactly they would use to identify these images as
AI. Figure~\ref{fig:hard-ai} shows all six images, each followed by two
regions of the image zoomed in to show details of artifacts identified by
our expert artists.

Next we summarize general techniques identified by experts, and use specific
images to illustrate these methods.

\begin{packed_itemize}
\item {\em Consistency in medium.} For any specific artwork, a trained artist
  typically employs only a single consistent medium, e.g. pencil, charcoal,
  and rarely combine multiple mediums.  For example, in the sketch in image
  a), our experts locate not only a ``weird halo effect'' (detail 1) due to the use of
  both pencil and charcoal, but also a ``crunchiness'' (detail 2) to the lines that
  associate with neither pencil nor charcoal.  AI models are associating
  lines from multiple mediums with the same style and fail to differentiate
  between them. Similarly, oil and acrylic paintings can display messy or
  smooth styles but never both. In image image d), the perfectly round moon
  looks like a digital art (detail 2), inconsistent with the overall messy
  painting style. The white is ``too clean'' (detail 1) when transitioning
  into the blue background.

\item {\em Intentionality in details.}  In art featuring human figures, human
  artists dedicate considerable effort to convey precise details of human
  features.  In image b), the light caught in the eyes do not match and the
  hair ends flow in opposite directions from the rest of the hair (detail 2).
  Similarly, in image c), hairs behind the neck are floating and doing
  completely different things from the rest of her hair.  Human artists also
  avoid unusual tangents with the bangs and eyebrow, something that this
  image completely overlooked.

\item {\em Limitations of medium.}  Experienced artists know that certain
  patterns and details are impossible to produce in real life due to physical
  limits of the medium. In image e), since watercolor is transparent and it
  bleeds after each brush stroke, the ``white over dark in the quills''
  (detail 1) is impossible to physically produce. The image is also too
  smooth to be hand painted on paper (detail 2), since watercolor bleeds in
  random directions.

\item {\em Domain knowledge.} There are specific rules when drawing specific
  subjects that are easy to validate. Wet paintings have an order of
  application, from light to dark, transparent to opaque. Thus all white
  spaces must be subtractive. Yet in image f), the white
  highlight is added after a dark area, which is wrong.  Also the water flow
  shows the wrong diffusion pattern. A trained artist should not make
  these mistakes.  \vspace{-0.05in}
   \end{packed_itemize}

 We consider these techniques in aggregate, and note that most of them
 require significant training and external knowledge to apply. In this sense,
 factors such as intentionality and domain knowledge seem like the most
 difficult for AI models to capture from training data. But as a whole, these
 domain-specific filters clearly operate very differently from statistical
 approaches used in automated detectors.  Thus we believe these methods will
 continue to be effective even as AI models continue to evolve.

 \subsection{Summary of Key Findings}
 Our multi-tier user study finds that general (non-artist) users are unable
 to distinguish between human art and AI-generated images. Professional
 artists are more confident and make more accurate decisions, and experienced
 experts are the most effective.  Diving deeper into concrete examples, we
 learn that experts leverage extensive knowledge of art mediums and
 techniques to identify what features are physically impossible, and 
 inconsistencies and mistakes that professional artists would avoid.

\begin{table}[t]
  \centering 
    \resizebox{0.48\textwidth}{!}{
    \centering
\begin{tabular}{r|ccc|ccc}
\multirow{2}{*}{Metric} & \multicolumn{3}{c|}{Human detector}                                                       & \multicolumn{3}{c}{Machine detector}                                                        \\ \cline{2-7} 
                            & \multicolumn{1}{c}{General} & \multicolumn{1}{c}{Professional} & \multicolumn{1}{c|}{Expert} & \multicolumn{1}{c}{Hive} & \multicolumn{1}{c}{Optic} & \multicolumn{1}{c}{Illuminarty} \\ \hline
ACC (\%) $\uparrow$                 & 59.23                       & 75.32                      & 83.00                       & 98.03                   & 90.67                     & 72.65                           \\
FPR (\%)  $\downarrow$  & 40.81                       & 23.53                      & 20.78                       & 0.00                     & 24.47                     & 67.40                           \\
FNR (\%) $\downarrow$ & 40.75                       & 25.37                      & 14.63                       & 3.17                     & 1.15                      & 4.69                 \\\hline         
\end{tabular}}
\caption{Performance of human and machine detectors on unperturbed imagery.}
\label{tab:human-ai-general-table}
\vspace{-0.2in}
\end{table}

\section{Human vs. Classifier Detectors} 
\label{sec:findings}

In this section, we present our findings by comparing the performance of human and machine detectors.  Our analysis starts from the baseline results, where all detectors are tested using the same set of usual images, covering unperturbed human artworks and AI-generated images.  Next, we investigate how human detectors respond to Glazed images, for which machine detectors have struggled to handle (as shown in \S\ref{sub:perturbresult}).  Finally, we explore both human and machine detectors' responses to unusual images,  including both hybrid images (when human edits AI-generated images) and upscaled human art (where artists polish the digital image of their artworks using enhancement tools). 

\subsection{Decision Accuracy and Confidence}
We start from comparing human and machine detectors on the baseline task of differentiating 

\para{Detection accuracy.} Table~\ref{tab:human-ai-general-table} lists the detection performance of both human detectors and classifier-based detectors, on unperturbed images.  Among the six detectors,  we have Hive $>$ Optic $>$ Expert Artist $>$ Professional Artist $>$ Illuminarty $>$ Non-Artist. 

\begin{figure}[t]
    \centering
    \includegraphics[width=0.4\textwidth]{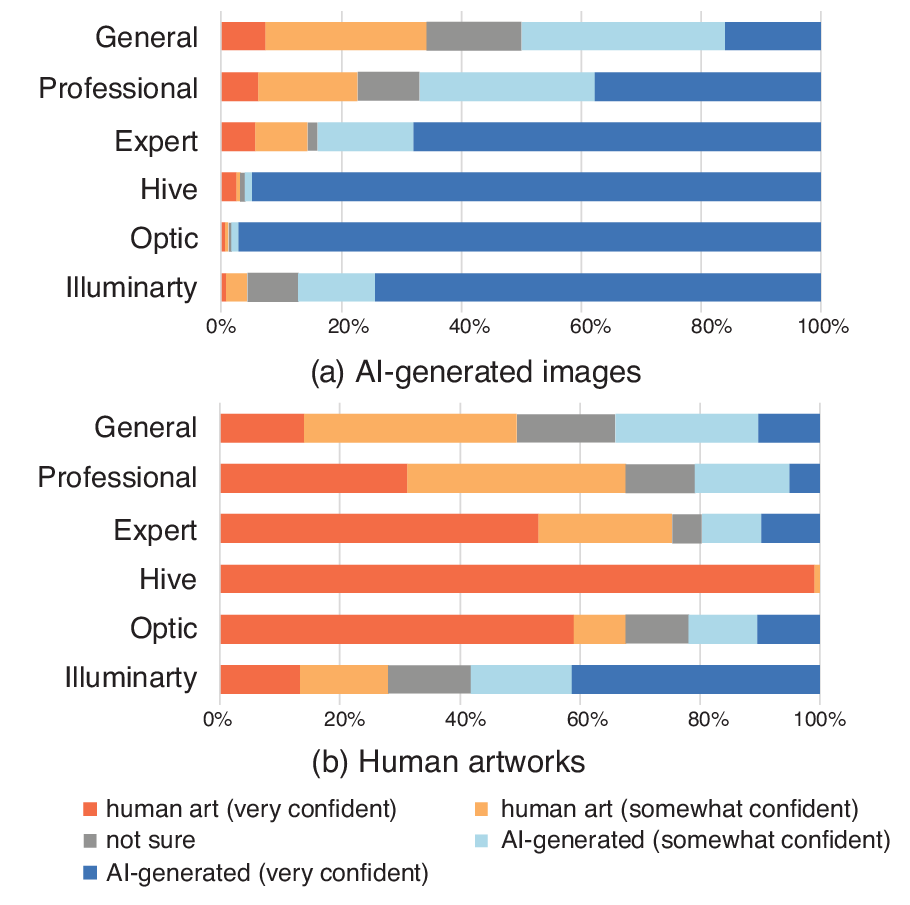}
    \vspace{-0.1in}
    \caption{Distribution of detection decision represented by the 5-point Likert rating on (a) AI-generated images and (b) human artworks.}\vspace{-0.2in}
    \label{fig:decision_confidence}
\end{figure}

\para{Decision confidence.}  We are also interested in understanding the distribution of decision confidence among human and machine detectors.   Figure~\ref{fig:decision_confidence} shows,  for AI-generated images and human artworks,  the distribution of the ``raw'' decision represented by the 5-point Likert score used by our study.  For AI-generated images (shown by the top figure),  the dark blue bar represents the ratio of correct decisions that also carry high confidence, while for human artworks (the bottom figure), the dark orange bar captures the ratio of correct decisions made with high confidence.

Across the three groups of human detectors,  general users are the least accuracy and also show the lowest confidence in their decisions, while expert artists are the most accurate and the most confident. This is as expected.  Across the three machine detectors, Hive is highly confident (and accurate), followed by Optic. Additionally,  Optic is more confident when facing  AI-generated images than human artworks.  Overall, the two machine detectors (Hive and Optic) show higher confidence (and higher accuracy) than expert artists, while the third machine detector (Illuminarty) performs worse than both expert and professional artists.

\subsection{Are Human Artists Better at Judging Glazed Images?}
\label{subsec:testglaze} 
As shown by Table~\ref{tab:glaze-table} in \S\ref{sub:perturbresult},  machine detectors, especially Hive and Optic, are much less effective at judging Glazed versions of AI-generated images.  Thus a natural question is ``would artists who know or use Glaze be more effective at judging Glazed images?''  To answer this question, we conduct an additional user study with our expert group, who all use Glaze on their published artworks.  Here we randomly select 100 AI-generated images (20 images per AI generator) and 100 human artworks (15 images per style, except 10 images for cartoon), and use the WebGlaze tool and the medium intensity setting to Glaze all 200 images.  For each expert artist, we randomly select 20 Glazed images and ask them to decide between human art and AI-generated.

Table~\ref{tab:glaze-med-userstudy} lists the overall performance across Glazed images. We see that human experts largely outperform both Optic and Illuminarty at judging Glazed images, whether it is human art or AI-generated. While Hive is still the best-performing detector in the overall accuracy (87.76\%),  our expert artists are not far off (83.44\%).  But more importantly,   human experts achieve a much lower FNR (8.97\%) compared to Hive (20.62\%).  This implies that {\bf human artists do outperform machines at judging Glazed versions of AI-generated images}.

\begin{table}[t]
    \centering
    \resizebox{0.36\textwidth}{!}{
    \centering
    \begin{tabular}{c|ccc}
      & ACC (\%)$\uparrow$ & FPR (\%) $\downarrow$ & FNR (\%) $\downarrow$ \\ \hline
      Expert Artist    & 83.44    & 23.53    & 8.97    \\ 
Hive        & 87.76    & 4.04     & 20.62    \\
Optic       & 61.20    & 38.30    & 39.33    \\
Illuminarty & 66.11    & 59.09    & 9.78    \\ \hline
\end{tabular}}
\caption{Detection performance on Glazed version of human artworks and AI-generated images.}
\vspace{-0.2in}
\label{tab:glaze-med-userstudy}
\end{table}

\subsection{Judging Unusual Images} 
\label{subsec:unusualresult}
As discussed in \S\ref{sec:unusualdata},  we also consider two unusual types of images: (i) hybrid images produced by human users editing AI-generated images, and (ii) upscaled human artworks where artists apply digital touchups to polish the image of their artworks. By evaluating them using both human and machine detectors, our goal is to identify, if any, notable differences in how they are evaluated by human and machine detectors.

In total, we collect 70 human artworks after applying upscaling and 40 hybrid images (see \S\ref{sec:unusualdata}).  Given the  limited member size of our expert group, we do not have sufficient coverage on these images by expert artists and thus omit their results. Figure~\ref{fig:unusual_decision_confidence} in Appendix plots, for both types of images, the distribution of the ``raw'' detection decision represented by the 5-point Likert score, for both human and machine detectors.  

\para{Hybrid Images.}  For these images, the decision distribution is similar to that of AI-generated images shown in Figure~\ref{fig:unusual_decision_confidence}(a), suggesting that both human and machine detectors frequently label these images as AI-generated.

\para{Upscaled Human Art.} The decision distribution is very similar to that of human artworks in Figure~\ref{fig:unusual_decision_confidence}(b).  This implies that upscaling does not have a significant impact of human artworks in terms of their decision outcomes from both human and machine detectors.

\secspace
\section{Combining Human and Automated Detectors}
\label{sec:combine}
Our study shows that both human artists and automated detectors face
challenges in distinguishing between human art and
AI-generated images. Tools like Hive are highly effective at
evaluating unperturbed images, but perform poorly when AI-generated
images are intentionally perturbed (e.g., Glaze or image overlays) to evade detection. On the other
hand, human experts can still identify perturbed
AI-generated images.  Thus we believe a mixed team of
human artists and machine classifiers will be the most effective.

\para{Teaming up Hive and Expert Artists.} We evaluate a scenario that
combines Hive scores with one expert artist. If Hive and the expert disagree, 
the score with higher confidence wins.

We evaluate the combined detector on both unperturbed artworks/images
and  their Glazed versions.  Table~\ref{tab:expert-hive-unperturbed-table} shows that, for
unperturbed images, the combined detector exhibits a slightly lower
accuracy compared to Hive.
Next, 
Table~\ref{tab:hive-1-expert-glazed-table} shows that the combined
detector is highly effective in judging Glazed images, outperforming
both Hive and expert.  Notably, it outperforms Hive by lowering FNR
from 19.70\% down to
8.82\%. Thus it is more effective
at identifying AI-generated images that have applied Glaze in an
attempt to evade detection.  At the same time, the combined detector
achieves a low FPR like Hive (6.06\%), remarkably lower than that of
expert (23.08\%).  Finally, the equilibrium between FPR and FNR
values, on 
both Glazed and original images
(Table~\ref{tab:expert-hive-unperturbed-table}), suggests ``unbiased''
detection accuracy for human artworks and AI-generated images.

\begin{table}[t]
  \centering 
    \resizebox{0.34\textwidth}{!}{
    \centering
    \begin{tabular}{c|ccc}
      \multicolumn{4}{c}{Unperturbed Human Artworks and AI Images} \\ \hline
      Detector                       & ACC (\%)$\uparrow$ & FPR (\%) $\downarrow$ & FNR (\%) $\downarrow$ \\ \hline
      Hive                     & 98.03              & 0.0                   & 3.17                  \\
      expert                  & 83.00              & 20.78                 & 14.63                 \\
      Hive + expert & 92.19              & 8.11                  & 7.63                  \\ \hline
    \end{tabular}}
  \caption{Detection performance on unperturbed human artworks and
    AI-generated images.  Three detectors: Hive, one expert (per
    image),  Hive + expert (tiebreak=confidence).}
  \label{tab:expert-hive-unperturbed-table}  \vspace{-0.2in}
\end{table}

\begin{table}[t]
  \centering 
    \resizebox{0.34\textwidth}{!}{
    \centering
    \begin{tabular}{c|ccc}
          \multicolumn{4}{c}{Glazed Human Artworks and AI Images} \\ \hline
      Detector                       & ACC (\%)$\uparrow$ & FPR (\%)
                                                            $\downarrow$
      & FNR (\%) $\downarrow$ \\ \hline
            Hive                     & 87.12              & 6.06                  & 19.70                 \\
      expert                    & 84.85              & 23.08                 & 7.46                  \\
      Hive + expert & 92.54              & 6.06                  & 8.82                  \\ \hline
    \end{tabular}}
  \caption{Detection performance on Glazed human artworks and
    AI-generated images.  Three detectors: one expert (per image),  Hive,  Hive + expert (tiebreak = confidence).}
  \label{tab:hive-1-expert-glazed-table}
  \vspace{-0.2in}
\end{table}

\section{Ethics}
\label{sec:ethics}

Our user study  was reviewed and approved by our institutional
review board (IRB). In our study, we prioritized consent and
protection of all participants, especially human artists and their
artworks.

\para{Consent for Human Art.} Our study necessitates the use of artwork by
human artists. To obtain consent, we identified artists and reached out with
request for permission. Many responded. We waited roughly 4 weeks, and
reached out again to everyone else. Once downloaded, images are anonymized
and stored on private, secure servers.

Since we asked for artwork from human artists with signatures/watermarks
removed, we were extremely sensitive to potential unauthorized exposure.
We took efforts to minimize the exposure of human artwork to external
sources. They were available to participating crowdsourced workers only for a
short time through the Prolific platform. In the regular artist user study,
images were available to participants for a total of 14 hours, after which we
shut down the study to minimize uncontrolled exposure. 

\para{Exposure to Web Services.} We took careful steps to ensure that images
of human art were not misused by external AI detection services. We reached
out to both Optic and Illuminarty, and were assured that images are never
used for training and deleted after process (with a max of 4 days for
Optic). Hive's terms of service states they can train AI models using images
uploaded via the free web service, but not images classified through paid
APIs. Thus we obtained access to a paid Hive account, and ensured all images
of human art were classified using this paid Hive account.

\secspace
\section{Discussion and Takeaways}
\label{sec:conclusion}

\annarebuttal{

As with any real-world study, there are limitations in our study that need to be considered. 
\begin{packed_itemize}
    \item Category of styles: our dataset only included a few fixed art styles. 
    More diverse styles might provide more comprehensive results.
    \item Cropping of images: cropping was applied to a small number of human artworks. 
    We avoided samples with highly irregular aspect ratios and ensured meticulous cropping to minimize irregularities.
    \item Curating Likert Scale: we discarded ``not sure'' responses from the user study. 
    This was done to maintain consistency with measuring the confidence level (40-60\%) in automated detectors and to prevent guesses from affecting our metrics.
\end{packed_itemize}

\noindent Our results also suggest takeaways for different audiences.

\para{For artists.} Proving human authenticity will be come increasingly
important, and increasingly difficult. No single method wil be foolproof, and
artists should consider incorporating work-in-progress (WIP) or timelapses
into their process.

\para{For researchers.} Developing highly accurate detection tools requires
continued investment in ethically obtaining diverse training sets. To enhance
robustness, incorporating perturbed and adversarially altered images during
training is crucial.

\para{For policy makers.} Implementing standards that mandate mixed detection
teams in critical applications can enhance both detection accuracy and
robustness.

As AI evolves, generative AI users seeking to avoid detection will adapt to
exploit vulnerabilities.  Some are using Adobe's Photoshop to add
imperfections to avoid the smooth AI finishing look.  Others are using one
model to generate foreground objects and another to create
backgrounds~\cite{lessai}.  These evolving techniques will continue to
present challenges for future AI detection systems.}

\subsection*{Acknowledgements}
We thank our anonymous reviewers for their insightful feedback. Sincere
thanks also go to the thousands of artists who participated in our user study. This work is
supported in part by NSF grants CNS-2241303 and CNS-1949650. Opinions, findings,
and conclusions or recommendations expressed in this material are those of
the authors and do not necessarily reflect the views of any funding agencies.

\bibliographystyle{ACM-Reference-Format}
\bibliography{detectai}


\begin{thebibliography}{80}


\ifx \showCODEN    \undefined \def \showCODEN     #1{\unskip}     \fi
\ifx \showDOI      \undefined \def \showDOI       #1{#1}\fi
\ifx \showISBNx    \undefined \def \showISBNx     #1{\unskip}     \fi
\ifx \showISBNxiii \undefined \def \showISBNxiii  #1{\unskip}     \fi
\ifx \showISSN     \undefined \def \showISSN      #1{\unskip}     \fi
\ifx \showLCCN     \undefined \def \showLCCN      #1{\unskip}     \fi
\ifx \shownote     \undefined \def \shownote      #1{#1}          \fi
\ifx \showarticletitle \undefined \def \showarticletitle #1{#1}   \fi
\ifx \showURL      \undefined \def \showURL       {\relax}        \fi
\providecommand\bibfield[2]{#2}
\providecommand\bibinfo[2]{#2}
\providecommand\natexlab[1]{#1}
\providecommand\showeprint[2][]{arXiv:#2}

\bibitem[AI(2023)]%
        {magnificai}
\bibfield{author}{\bibinfo{person}{Magnific AI}.}
  \bibinfo{year}{2023}\natexlab{}.
\newblock \bibinfo{title}{{Magnific}}.
\newblock
\newblock
\newblock
\shownote{\url{https://magnific.ai}}.


\bibitem[Alemohammad et~al\mbox{.}(2023)]%
        {genai-mad}
\bibfield{author}{\bibinfo{person}{Sina Alemohammad} {et~al\mbox{.}}}
  \bibinfo{year}{2023}\natexlab{}.
\newblock \showarticletitle{Self-Consuming Generative Models Go {MAD}}. In
  \bibinfo{booktitle}{\emph{arXiv preprint:2307.01850}}.
\newblock


\bibitem[BAIO(2022)]%
        {hollie-steal}
\bibfield{author}{\bibinfo{person}{ANDY BAIO}.}
  \bibinfo{year}{2022}\natexlab{}.
\newblock \bibinfo{title}{{Invasive Diffusion: How one unwilling illustrator
  found herself turned into an AI model}}.
\newblock
\newblock


\bibitem[Bammey(2020)]%
        {bammey2023synthbuster}
\bibfield{author}{\bibinfo{person}{Quentin Bammey}.}
  \bibinfo{year}{2020}\natexlab{}.
\newblock \showarticletitle{Synthbuster: Towards Detection of Diffusion Model
  Generated Images}.
\newblock \bibinfo{journal}{\emph{IEEE Open Journal of Signal Processing}}
  \bibinfo{volume}{5} (\bibinfo{year}{2020}), \bibinfo{pages}{1--9}.
\newblock


\bibitem[Bi et~al\mbox{.}(2023)]%
        {bi2023detecting}
\bibfield{author}{\bibinfo{person}{Xiuli Bi} {et~al\mbox{.}}}
  \bibinfo{year}{2023}\natexlab{}.
\newblock \showarticletitle{Detecting Generated Images by Real Images Only}.
\newblock \bibinfo{journal}{\emph{arXiv preprint:2311.00962}}
  (\bibinfo{year}{2023}).
\newblock


\bibitem[Bird and Lotfi(2023)]%
        {bird2023cifake}
\bibfield{author}{\bibinfo{person}{Jordan~J. Bird} {and} \bibinfo{person}{Ahmad
  Lotfi}.} \bibinfo{year}{2023}\natexlab{}.
\newblock \showarticletitle{CIFAKE: Image Classification and Explainable
  Identification of AI-Generated Synthetic Images}.
\newblock \bibinfo{journal}{\emph{arXiv preprint:2303.14126}}
  (\bibinfo{year}{2023}).
\newblock


\bibitem[Bousquette(2023)]%
        {fakeart-backlash}
\bibfield{author}{\bibinfo{person}{Isabelle Bousquette}.}
  \bibinfo{year}{2023}\natexlab{}.
\newblock \bibinfo{title}{Companies Increasingly Fear Backlash Over Their {AI}
  Work}.
\newblock \bibinfo{howpublished}{WSJ}.
\newblock


\bibitem[Braudaway(1997)]%
        {braudaway1997protecting}
\bibfield{author}{\bibinfo{person}{G.~W. Braudaway}.}
  \bibinfo{year}{1997}\natexlab{}.
\newblock \showarticletitle{Protecting publicly-available images with an
  invisible image watermark}. In \bibinfo{booktitle}{\emph{Proc. of ICIP}}.
\newblock


\bibitem[Bray et~al\mbox{.}(2023)]%
        {bray2023testing}
\bibfield{author}{\bibinfo{person}{Sergi~D. Bray} {et~al\mbox{.}}}
  \bibinfo{year}{2023}\natexlab{}.
\newblock \showarticletitle{Testing human ability to detect `deepfake' images
  of human faces}.
\newblock \bibinfo{journal}{\emph{Journal of Cybersecurity}}
  (\bibinfo{year}{2023}), \bibinfo{pages}{1--18}.
\newblock


\bibitem[Briesch et~al\mbox{.}(2023)]%
        {genai-mad2}
\bibfield{author}{\bibinfo{person}{Martin Briesch}, \bibinfo{person}{Dominik
  Sobania}, {and} \bibinfo{person}{Franz Rothlauf}.}
  \bibinfo{year}{2023}\natexlab{}.
\newblock \showarticletitle{Large Language Models Suffer From Their Own Output:
  An Analysis of the Self-Consuming Training Loop}. In
  \bibinfo{booktitle}{\emph{arXiv preprint:2311.16822}}.
\newblock


\bibitem[Cara(2023)]%
        {cara}
\bibfield{author}{\bibinfo{person}{Cara}.} \bibinfo{year}{2023}\natexlab{}.
\newblock \bibinfo{howpublished}{\url{https://cara.app/}}.
\newblock


\bibitem[Chyung et~al\mbox{.}(2017)]%
        {chyung2017evidence}
\bibfield{author}{\bibinfo{person}{Seung~Youn Chyung} {et~al\mbox{.}}}
  \bibinfo{year}{2017}\natexlab{}.
\newblock \showarticletitle{Evidence-Based Survey Design: The Use of a Midpoint
  on the Likert Scale}.
\newblock \bibinfo{journal}{\emph{Performance Improvement}}
  (\bibinfo{year}{2017}), \bibinfo{pages}{15--23}.
\newblock


\bibitem[{Civitai}(2022)]%
        {civitai}
\bibfield{author}{\bibinfo{person}{{Civitai}}.}
  \bibinfo{year}{2022}\natexlab{}.
\newblock \bibinfo{title}{{What is Civitai?}}
\newblock
\newblock
\newblock
\shownote{\url{https://civitai.com/content/guides/what-is-civitai}}.


\bibitem[Codega(2023a)]%
        {fakeart-dnd2}
\bibfield{author}{\bibinfo{person}{Linda Codega}.}
  \bibinfo{year}{2023}\natexlab{a}.
\newblock \bibinfo{title}{Dungeons \& Dragons Updates Bigby to Replace
  {AI}-Enhanced Images}.
\newblock \bibinfo{howpublished}{Gizmodo}.
\newblock


\bibitem[Codega(2023b)]%
        {fakeart-dnd}
\bibfield{author}{\bibinfo{person}{Linda Codega}.}
  \bibinfo{year}{2023}\natexlab{b}.
\newblock \bibinfo{title}{New Dungeons \& Dragons Sourcebook Features {AI}
  Generated Art}.
\newblock \bibinfo{howpublished}{Gizmodo}.
\newblock


\bibitem[Corvi et~al\mbox{.}(2023)]%
        {corvi2023detection}
\bibfield{author}{\bibinfo{person}{Riccardo Corvi} {et~al\mbox{.}}}
  \bibinfo{year}{2023}\natexlab{}.
\newblock \showarticletitle{On The Detection of Synthetic Images Generated by
  Diffusion Models}. In \bibinfo{booktitle}{\emph{Proc. of ICASSP}}.
\newblock


\bibitem[CourtListener(2024)]%
        {sdlitigation}
\bibfield{author}{\bibinfo{person}{CourtListener}.}
  \bibinfo{year}{2024}\natexlab{}.
\newblock \bibinfo{title}{{Andersen v. Stability AI Ltd.(3:23-cv-00201)}}.
\newblock
\newblock
\newblock
\shownote{\url{https://www.courtlistener.com/docket/66732129/andersen-v-stability-ai-ltd/}}.


\bibitem[Cozzolino et~al\mbox{.}(2023)]%
        {cozzolino2023raising}
\bibfield{author}{\bibinfo{person}{Davide Cozzolino} {et~al\mbox{.}}}
  \bibinfo{year}{2023}\natexlab{}.
\newblock \showarticletitle{Raising the Bar of AI-generated Image Detection
  with CLIP}.
\newblock \bibinfo{journal}{\emph{arXiv preprint:2312.00195}}
  (\bibinfo{year}{2023}).
\newblock


\bibitem[Deighton(2024)]%
        {lessai}
\bibfield{author}{\bibinfo{person}{Katie Deighton}.}
  \bibinfo{year}{2024}\natexlab{}.
\newblock \bibinfo{title}{How the Ad Industry Is Making AI Images Look Less
  Like AI}.
\newblock \bibinfo{howpublished}{Wall Street Journal}.
\newblock


\bibitem[Demontis et~al\mbox{.}(2019)]%
        {demontis2019adversarial}
\bibfield{author}{\bibinfo{person}{Ambra Demontis} {et~al\mbox{.}}}
  \bibinfo{year}{2019}\natexlab{}.
\newblock \showarticletitle{Why do adversarial attacks transfer? explaining
  transferability of evasion and poisoning attacks}. In
  \bibinfo{booktitle}{\emph{Proc. of {USENIX} Security}}.
\newblock


\bibitem[Dhariwal and Nichol(2021)]%
        {dhariwal2021diffusion}
\bibfield{author}{\bibinfo{person}{Prafulla Dhariwal} {and}
  \bibinfo{person}{Alexander Nichol}.} \bibinfo{year}{2021}\natexlab{}.
\newblock \showarticletitle{Diffusion Models Beat GANs on Image Synthesis}. In
  \bibinfo{booktitle}{\emph{Proc. of {NeurIPS}}}.
\newblock


\bibitem[Dupre(2023)]%
        {fakeart-backlash2}
\bibfield{author}{\bibinfo{person}{Maggie~H. Dupre}.}
  \bibinfo{year}{2023}\natexlab{}.
\newblock \bibinfo{title}{Sports Illustrated Publisher Fires {CEO} After {AI}
  Scandal}.
\newblock \bibinfo{howpublished}{Futurism}.
\newblock


\bibitem[Emmanuel(2023)]%
        {newsstory1}
\bibfield{author}{\bibinfo{person}{Salako Emmanuel}.}
  \bibinfo{year}{2023}\natexlab{}.
\newblock \bibinfo{title}{{AI Tools for Combating Deepfakes.}}
\newblock
\newblock
\newblock
\shownote{\url{https://ijnet.org/en/story/ai-tools-combating-deepfakes}}.


\bibitem[Frank et~al\mbox{.}(2024)]%
        {cispa-detect}
\bibfield{author}{\bibinfo{person}{Joel Frank} {et~al\mbox{.}}}
  \bibinfo{year}{2024}\natexlab{}.
\newblock \showarticletitle{A Representative Study on Human Detection of
  Artificially Generated Media Across Countries}. In
  \bibinfo{booktitle}{\emph{Proc. of IEEE S\&P}}. \bibinfo{address}{San
  Francisco, CA}.
\newblock


\bibitem[Gach(2023)]%
        {fakeart-amazon}
\bibfield{author}{\bibinfo{person}{Ethan Gach}.}
  \bibinfo{year}{2023}\natexlab{}.
\newblock \bibinfo{title}{Amazon's First Official Fallout TV Show Artwork Is an
  {AI}-Looking Eyesore}.
\newblock \bibinfo{howpublished}{Kotaku.com}.
\newblock


\bibitem[Ghazanfari et~al\mbox{.}(2023)]%
        {ghazanfari2023r}
\bibfield{author}{\bibinfo{person}{Sara Ghazanfari} {et~al\mbox{.}}}
  \bibinfo{year}{2023}\natexlab{}.
\newblock \showarticletitle{R-LPIPS: An adversarially robust perceptual
  similarity metric}.
\newblock \bibinfo{journal}{\emph{arXiv preprint:2307.15157}}
  (\bibinfo{year}{2023}).
\newblock


\bibitem[Glynn(2023)]%
        {photographycomp}
\bibfield{author}{\bibinfo{person}{Paul Glynn}.}
  \bibinfo{year}{2023}\natexlab{}.
\newblock \bibinfo{title}{{Sony World Photography Award 2023: Winner Refuses
  Award After Revealing AI Creation.}}
\newblock
\newblock
\newblock
\shownote{BBC News}.


\bibitem[Goodfellow et~al\mbox{.}(2014)]%
        {goodfellow2014explaining}
\bibfield{author}{\bibinfo{person}{Ian~J Goodfellow}, \bibinfo{person}{Jonathon
  Shlens}, {and} \bibinfo{person}{Christian Szegedy}.}
  \bibinfo{year}{2014}\natexlab{}.
\newblock \showarticletitle{Explaining and harnessing adversarial examples}.
\newblock \bibinfo{journal}{\emph{arXiv preprint:1412.6572}}
  (\bibinfo{year}{2014}).
\newblock


\bibitem[Hamano et~al\mbox{.}(2023)]%
        {hamano2023effects}
\bibfield{author}{\bibinfo{person}{Genki Hamano}, \bibinfo{person}{Shoko
  Imaizumi}, {and} \bibinfo{person}{Hitoshi Kiya}.}
  \bibinfo{year}{2023}\natexlab{}.
\newblock \showarticletitle{Effects of JPEG Compression on Vision Transformer
  Image Classification for Encryption-then-Compression Images}.
\newblock \bibinfo{journal}{\emph{Sensors}} \bibinfo{volume}{23},
  \bibinfo{number}{7} (\bibinfo{year}{2023}).
\newblock


\bibitem[Hive(2023)]%
        {hivedetection}
\bibfield{author}{\bibinfo{person}{Hive}.} \bibinfo{year}{2023}\natexlab{}.
\newblock \bibinfo{title}{{AI-Generated Content Classification}}.
\newblock
\newblock
\newblock
\shownote{\url{https://thehive.ai/apis/ai-generated-content-classification}}.


\bibitem[Ho(2024)]%
        {mid-list}
\bibfield{author}{\bibinfo{person}{Karen Ho}.} \bibinfo{year}{2024}\natexlab{}.
\newblock \bibinfo{title}{{Database of 16,000 Artists Used to Train Midjourney
  AI, Including 6-Year-Old Child, Garners Criticism}}.
\newblock
\newblock
\newblock
\shownote{\url{https://www.artnews.com/art-news/news/midjourney-ai-artists-database-1234691955/}}.


\bibitem[Hooda et~al\mbox{.}(2024)]%
        {hooda2024d4}
\bibfield{author}{\bibinfo{person}{Ashish Hooda} {et~al\mbox{.}}}
  \bibinfo{year}{2024}\natexlab{}.
\newblock \showarticletitle{D4: Detection of Adversarial Diffusion Deepfakes
  Using Disjoint Ensembles}. In \bibinfo{booktitle}{\emph{Proc. of {WACV}}}.
  IEEE.
\newblock


\bibitem[Illuminarty(2023)]%
        {illuminarty}
\bibfield{author}{\bibinfo{person}{Illuminarty}.}
  \bibinfo{year}{2023}\natexlab{}.
\newblock \bibinfo{title}{{Is an AI Behind Your Image?}}
\newblock
\newblock
\newblock
\shownote{\url{https://illuminarty.ai/}}.


\bibitem[Jeong and Lee(2016)]%
        {jeong2016level}
\bibfield{author}{\bibinfo{person}{Heon~Jae Jeong} {and}
  \bibinfo{person}{Wui~Chiang Lee}.} \bibinfo{year}{2016}\natexlab{}.
\newblock \showarticletitle{The level of collapse we are allowed: comparison of
  different response scales in safety attitudes questionnaire}.
\newblock \bibinfo{journal}{\emph{Biometrics Biostatistics International
  Journal}} (\bibinfo{year}{2016}), \bibinfo{pages}{128--134}.
\newblock


\bibitem[{Joseph Saveri Law Firm LLP}(2023)]%
        {class-action}
\bibfield{author}{\bibinfo{person}{{Joseph Saveri Law Firm LLP}}.}
  \bibinfo{year}{2023}\natexlab{}.
\newblock \bibinfo{title}{{Class Action Filed Against Stability AI, Midjourney,
  and DeviantArt for DMCA Violations, Right of Publicity Violations, Unlawful
  Competition, Breach of TOS}}.
\newblock
\newblock


\bibitem[Karras et~al\mbox{.}(2020)]%
        {karras2020analyzing}
\bibfield{author}{\bibinfo{person}{Tero Karras} {et~al\mbox{.}}}
  \bibinfo{year}{2020}\natexlab{}.
\newblock \showarticletitle{Analyzing and Improving the Image Quality of
  StyleGAN}.
\newblock \bibinfo{journal}{\emph{arXiv preprint:1912.04958}}
  (\bibinfo{year}{2020}).
\newblock


\bibitem[Krizhevsky and Hinton(2009)]%
        {krizhevsky2009cifar}
\bibfield{author}{\bibinfo{person}{Alex Krizhevsky} {and}
  \bibinfo{person}{Geoffrey Hinton}.} \bibinfo{year}{2009}\natexlab{}.
\newblock \showarticletitle{Learning multiple layers of features from tiny
  images}.
\newblock \bibinfo{journal}{\emph{University of Toronto}}
  (\bibinfo{year}{2009}).
\newblock


\bibitem[Lab(2023)]%
        {webglaze}
\bibfield{author}{\bibinfo{person}{SAND Lab}.} \bibinfo{year}{2023}\natexlab{}.
\newblock \bibinfo{title}{{Web Glaze}}.
\newblock
\newblock
\newblock
\shownote{\url{https://glaze.cs.uchicago.edu/webglaze.html}}.


\bibitem[Lam et~al\mbox{.}(1999)]%
        {lam1999effects}
\bibfield{author}{\bibinfo{person}{K.~W.~K. Lam}, \bibinfo{person}{W.~L. Lau},
  {and} \bibinfo{person}{Z.~L. Li}.} \bibinfo{year}{1999}\natexlab{}.
\newblock \showarticletitle{Effects of JPEG compression on accuracy of image
  classification}. In \bibinfo{booktitle}{\emph{Proc. of ACRS}}.
\newblock


\bibitem[Lau et~al\mbox{.}(2003)]%
        {lau2003effects}
\bibfield{author}{\bibinfo{person}{W-L Lau}, \bibinfo{person}{Z-L Li}, {and}
  \bibinfo{person}{KW-K Lam}.} \bibinfo{year}{2003}\natexlab{}.
\newblock \showarticletitle{Effects of JPEG compression on image
  classification}.
\newblock \bibinfo{journal}{\emph{Proc. of IJRS}} (\bibinfo{year}{2003}).
\newblock


\bibitem[Li et~al\mbox{.}(2023)]%
        {li2023blip}
\bibfield{author}{\bibinfo{person}{Junnan Li} {et~al\mbox{.}}}
  \bibinfo{year}{2023}\natexlab{}.
\newblock \showarticletitle{BLIP-2: bootstrapping language-image pre-training
  with frozen image encoders and large language models}. In
  \bibinfo{booktitle}{\emph{Proc. of {ICML}}}.
\newblock


\bibitem[Lu et~al\mbox{.}(2023)]%
        {lu2023seeing}
\bibfield{author}{\bibinfo{person}{Zeyu Lu} {et~al\mbox{.}}}
  \bibinfo{year}{2023}\natexlab{}.
\newblock \showarticletitle{Seeing is not always believing: Benchmarking Human
  and Model Perception of AI-Generated Images}.
\newblock \bibinfo{journal}{\emph{arXiv preprint:2304.13023}}
  (\bibinfo{year}{2023}).
\newblock


\bibitem[Madry et~al\mbox{.}(2017)]%
        {madry2017towards}
\bibfield{author}{\bibinfo{person}{Aleksander Madry} {et~al\mbox{.}}}
  \bibinfo{year}{2017}\natexlab{}.
\newblock \showarticletitle{Towards deep learning models resistant to
  adversarial attacks}.
\newblock \bibinfo{journal}{\emph{arXiv preprint:1706.06083}}
  (\bibinfo{year}{2017}).
\newblock


\bibitem[Maiberg(2023)]%
        {detectors-abuse}
\bibfield{author}{\bibinfo{person}{Emanuel Maiberg}.}
  \bibinfo{year}{2023}\natexlab{}.
\newblock \bibinfo{title}{{AI} Images Detectors Are Being Used to Discredit the
  Real Horrors of War}.
\newblock \bibinfo{howpublished}{404Media}.
\newblock


\bibitem[MaxonVFX(2024)]%
        {maxon-sorry}
\bibfield{author}{\bibinfo{person}{MaxonVFX}.} \bibinfo{year}{2024}\natexlab{}.
\newblock \bibinfo{title}{We extend our apologies to the community}.
\newblock
  \bibinfo{howpublished}{\url{https://twitter.com/MaxonVFX/status/1748826148858208286}}.
\newblock


\bibitem[Nichol et~al\mbox{.}(2022)]%
        {nichol2022hierarchical}
\bibfield{author}{\bibinfo{person}{Alex Nichol} {et~al\mbox{.}}}
  \bibinfo{year}{2022}\natexlab{}.
\newblock \showarticletitle{Hierarchical text-conditional image generation with
  clip latents}.
\newblock \bibinfo{journal}{\emph{arXiv preprint:2204.06125}}
  (\bibinfo{year}{2022}).
\newblock


\bibitem[Northup(2023)]%
        {wotc-sorry}
\bibfield{author}{\bibinfo{person}{Travis Northup}.}
  \bibinfo{year}{2023}\natexlab{}.
\newblock \bibinfo{title}{Wizards of the Coast Repeats Anti-{AI} Art Stance
  After Player's Handbook Controversy}.
\newblock \bibinfo{howpublished}{IGN.com}.
\newblock


\bibitem[Ong(2023)]%
        {fakeart-daphne}
\bibfield{author}{\bibinfo{person}{Jie~Yee Ong}.}
  \bibinfo{year}{2023}\natexlab{}.
\newblock \bibinfo{title}{Scooby-Doo: Daphne Voice Actor Fell Victim To \$1,000
  {AI} Art Scam}.
\newblock \bibinfo{howpublished}{The Chainsaw}.
\newblock


\bibitem[Optic(2023)]%
        {aiornot}
\bibfield{author}{\bibinfo{person}{Optic}.} \bibinfo{year}{2023}\natexlab{}.
\newblock \bibinfo{title}{{AI or Not}}.
\newblock
\newblock
\newblock
\shownote{\url{https://www.aiornot.com}}.


\bibitem[Orland(2024)]%
        {magicthegatheringcontroversy}
\bibfield{author}{\bibinfo{person}{Kyle Orland}.}
  \bibinfo{year}{2024}\natexlab{}.
\newblock \bibinfo{title}{{Magic: The Gathering Maker Admits it Used
  AI-generated Art Despite Standing Ban}}.
\newblock
\newblock
\newblock
\shownote{Ars Technica}.


\bibitem[Page-Katz(2023)]%
        {kickstarterai-policy}
\bibfield{author}{\bibinfo{person}{Susannah Page-Katz}.}
  \bibinfo{year}{2023}\natexlab{}.
\newblock \bibinfo{title}{Introducing Our AI Policy}.
\newblock \bibinfo{howpublished}{Kickstarter.com}.
\newblock


\bibitem[Podell et~al\mbox{.}(2023)]%
        {podell2023sdxl}
\bibfield{author}{\bibinfo{person}{Dustin Podell} {et~al\mbox{.}}}
  \bibinfo{year}{2023}\natexlab{}.
\newblock \showarticletitle{SDXL: Improving Latent Diffusion Models for
  High-Resolution Image Synthesis}.
\newblock \bibinfo{journal}{\emph{arXiv preprint:2307.01952}}
  (\bibinfo{year}{2023}).
\newblock


\bibitem[Radford et~al\mbox{.}(2021)]%
        {radford2021learning}
\bibfield{author}{\bibinfo{person}{Alec Radford} {et~al\mbox{.}}}
  \bibinfo{year}{2021}\natexlab{}.
\newblock \showarticletitle{Learning transferable visual models from natural
  language supervision}. In \bibinfo{booktitle}{\emph{Proc. of {ICML}}}.
\newblock


\bibitem[Ramesh et~al\mbox{.}(2022)]%
        {ramesh2022hierarchical}
\bibfield{author}{\bibinfo{person}{Aditya Ramesh} {et~al\mbox{.}}}
  \bibinfo{year}{2022}\natexlab{}.
\newblock \showarticletitle{Hierarchical text-conditional image generation with
  clip latents}.
\newblock \bibinfo{journal}{\emph{arXiv preprint:2204.06125}}
  (\bibinfo{year}{2022}).
\newblock


\bibitem[Ricker et~al\mbox{.}(2023)]%
        {ricker2023towards}
\bibfield{author}{\bibinfo{person}{Jonas Ricker}, \bibinfo{person}{Simon Damm},
  \bibinfo{person}{Thorsten Holz}, {and} \bibinfo{person}{Asja Fischer}.}
  \bibinfo{year}{2023}\natexlab{}.
\newblock \showarticletitle{Towards the Detection of Diffusion Model
  Deepfakes}.
\newblock \bibinfo{journal}{\emph{arXiv preprint:2210.14571}}
  (\bibinfo{year}{2023}).
\newblock


\bibitem[Rombach et~al\mbox{.}(2022)]%
        {rombach2022high}
\bibfield{author}{\bibinfo{person}{Robin Rombach} {et~al\mbox{.}}}
  \bibinfo{year}{2022}\natexlab{}.
\newblock \showarticletitle{High-resolution image synthesis with latent
  diffusion models}. In \bibinfo{booktitle}{\emph{Proc. of {CVPR}}}.
\newblock


\bibitem[Roose(2022)]%
        {winaward}
\bibfield{author}{\bibinfo{person}{Kevin Roose}.}
  \bibinfo{year}{2022}\natexlab{}.
\newblock \bibinfo{title}{{An A.I.-Generated Picture Won an Art Prize. Artists
  Aren't Happy.}}
\newblock
\newblock
\newblock
\shownote{\url{https://www.nytimes.com/2022/09/02/technology/ai-artificial-intelligence-artists.html}}.


\bibitem[Sato(2023)]%
        {fakeart-contest}
\bibfield{author}{\bibinfo{person}{Mia Sato}.} \bibinfo{year}{2023}\natexlab{}.
\newblock \bibinfo{title}{How {AI} art killed an indie book cover contest}.
\newblock \bibinfo{howpublished}{The Verge}.
\newblock


\bibitem[Schuhmann et~al\mbox{.}(2022)]%
        {schuhmann2022laion}
\bibfield{author}{\bibinfo{person}{Christoph Schuhmann} {et~al\mbox{.}}}
  \bibinfo{year}{2022}\natexlab{}.
\newblock \showarticletitle{Laion-5b: An open large-scale dataset for training
  next generation image-text models}.
\newblock \bibinfo{journal}{\emph{arXiv preprint:2210.08402}}
  (\bibinfo{year}{2022}).
\newblock


\bibitem[Sha et~al\mbox{.}(2023)]%
        {sha2023defake}
\bibfield{author}{\bibinfo{person}{Zeyang Sha}, \bibinfo{person}{Zheng Li},
  \bibinfo{person}{Ning Yu}, {and} \bibinfo{person}{Yang Zhang}.}
  \bibinfo{year}{2023}\natexlab{}.
\newblock \showarticletitle{DE-FAKE: Detection and Attribution of Fake Images
  Generated by Text-to-Image Generation Models}. In
  \bibinfo{booktitle}{\emph{Proc. of {CCS}}}.
\newblock


\bibitem[Shan et~al\mbox{.}(2023)]%
        {shan2023glaze}
\bibfield{author}{\bibinfo{person}{Shawn Shan}, \bibinfo{person}{Jenna Cryan},
  \bibinfo{person}{Emily Wenger}, \bibinfo{person}{Haitao Zheng},
  \bibinfo{person}{Rana Hanocka}, {and} \bibinfo{person}{Ben~Y. Zhao}.}
  \bibinfo{year}{2023}\natexlab{}.
\newblock \showarticletitle{Glaze: Protecting artists from style mimicry by
  text-to-image models}. In \bibinfo{booktitle}{\emph{Proc. of {USENIX}
  Security}}.
\newblock


\bibitem[Show"(2024)]%
        {fakeart-eons}
\bibfield{author}{\bibinfo{person}{"Eons Show"}.}
  \bibinfo{year}{2024}\natexlab{}.
\newblock \bibinfo{title}{Eons Show apologies for violating own {AI} policy}.
\newblock \bibinfo{howpublished}{Twitter}.
\newblock
\newblock
\shownote{\url{https://x.com/EonsShow/status/1751327424556544451}}.


\bibitem[Shumailov et~al\mbox{.}(2023)]%
        {shumailov2023curse}
\bibfield{author}{\bibinfo{person}{Ilia Shumailov} {et~al\mbox{.}}}
  \bibinfo{year}{2023}\natexlab{}.
\newblock \showarticletitle{The Curse of Recursion: Training on Generated Data
  Makes Models Forget}.
\newblock \bibinfo{journal}{\emph{arXiv preprint:2305.17493}}
  (\bibinfo{year}{2023}).
\newblock


\bibitem[Small(2023)]%
        {fakeart-copyright}
\bibfield{author}{\bibinfo{person}{Zachary Small}.}
  \bibinfo{year}{2023}\natexlab{}.
\newblock \bibinfo{title}{As Fight Over A.I. Artwork Unfolds, Judge Rejects
  Copyright Claim}.
\newblock \bibinfo{howpublished}{NY Times}.
\newblock


\bibitem[Song et~al\mbox{.}(2023)]%
        {song2023robustness}
\bibfield{author}{\bibinfo{person}{Haixu Song}, \bibinfo{person}{Shiyu Huang},
  \bibinfo{person}{Yinpeng Dong}, {and} \bibinfo{person}{Wei-Wei Tu}.}
  \bibinfo{year}{2023}\natexlab{}.
\newblock \showarticletitle{Robustness and Generalizability of Deepfake
  Detection: A Study with Diffusion Models}.
\newblock \bibinfo{journal}{\emph{arXiv preprint:2309.02218}}
  (\bibinfo{year}{2023}).
\newblock


\bibitem[Spring et~al\mbox{.}(2016)]%
        {spring2016jpeg}
\bibfield{author}{\bibinfo{person}{Kenneth~R. Spring}, \bibinfo{person}{John~C.
  Russ}, \bibinfo{person}{Matthew~J. Parry-Hill}, {and}
  \bibinfo{person}{Michael~W. Davidson}.} \bibinfo{year}{2016}\natexlab{}.
\newblock \bibinfo{title}{JPEG Image Compression}.
\newblock \bibinfo{howpublished}{National High Magnetic Field Laboratory}.
\newblock


\bibitem[{Stability AI}(2022)]%
        {sd-release}
\bibfield{author}{\bibinfo{person}{{Stability AI}}.}
  \bibinfo{year}{2022}\natexlab{}.
\newblock \bibinfo{title}{{Stable Diffusion Public Release}}.
\newblock
\newblock
\newblock
\shownote{\url{https://stability.ai/blog/stable-diffusion-public-release}}.


\bibitem[{StabilityAI}(2022a)]%
        {sd-14-modelcard}
\bibfield{author}{\bibinfo{person}{{StabilityAI}}.}
  \bibinfo{year}{2022}\natexlab{a}.
\newblock \bibinfo{title}{{Stable Diffusion v1-4 Model Card}}.
\newblock
\newblock
\newblock
\shownote{\url{https://huggingface.co/CompVis/stable-diffusion-v1-4}}.


\bibitem[{StabilityAI}(2022b)]%
        {sd-15-modelcard}
\bibfield{author}{\bibinfo{person}{{StabilityAI}}.}
  \bibinfo{year}{2022}\natexlab{b}.
\newblock \bibinfo{title}{{Stable Diffusion v1-5 Model Card}}.
\newblock
\newblock
\newblock
\shownote{\url{https://huggingface.co/runwayml/stable-diffusion-v1-5}}.


\bibitem[Staff"(2023)]%
        {riotlol-arttakedown}
\bibfield{author}{\bibinfo{person}{"Portal Staff"}.}
  \bibinfo{year}{2023}\natexlab{}.
\newblock \bibinfo{title}{League of Legends AI-Generated LATAM Anniversary
  Video Gets Taken Down}.
\newblock \bibinfo{howpublished}{ZLeague The Portal}.
\newblock


\bibitem[Steele(2023)]%
        {newsstory3}
\bibfield{author}{\bibinfo{person}{Chandra Steele}.}
  \bibinfo{year}{2023}\natexlab{}.
\newblock \bibinfo{title}{{How to Detect AI-Created Images}}.
\newblock
\newblock
\newblock
\shownote{\url{https://www.pcmag.com/how-to/how-to-detect-ai-created-images}}.


\bibitem[Thompson and Hsu(2023)]%
        {newsstory2}
\bibfield{author}{\bibinfo{person}{Stuart~A. Thompson} {and}
  \bibinfo{person}{Tiffany Hsu}.} \bibinfo{year}{2023}\natexlab{}.
\newblock \bibinfo{title}{How Easy Is It to Fool {A.I.}-Detection Tools?}
\newblock \bibinfo{howpublished}{NY Times}.
\newblock


\bibitem[Wang et~al\mbox{.}(2020)]%
        {wang2020cnngenerated}
\bibfield{author}{\bibinfo{person}{Sheng-Yu Wang} {et~al\mbox{.}}}
  \bibinfo{year}{2020}\natexlab{}.
\newblock \showarticletitle{CNN-generated images are surprisingly easy to
  spot... for now}.
\newblock \bibinfo{journal}{\emph{arXiv preprint:1912.11035}}
  (\bibinfo{year}{2020}).
\newblock


\bibitem[Wang et~al\mbox{.}(2023b)]%
        {wang2023benchmarking}
\bibfield{author}{\bibinfo{person}{Yabin Wang}, \bibinfo{person}{Zhiwu Huang},
  {and} \bibinfo{person}{Xiaopeng Hong}.} \bibinfo{year}{2023}\natexlab{b}.
\newblock \showarticletitle{Benchmarking Deepart Detection}.
\newblock \bibinfo{journal}{\emph{arXiv preprint:2302.14475}}
  (\bibinfo{year}{2023}).
\newblock


\bibitem[Wang et~al\mbox{.}(2023a)]%
        {Wang_2023_ICCV}
\bibfield{author}{\bibinfo{person}{Zhendong Wang} {et~al\mbox{.}}}
  \bibinfo{year}{2023}\natexlab{a}.
\newblock \showarticletitle{DIRE for Diffusion-Generated Image Detection}. In
  \bibinfo{booktitle}{\emph{Proceedings of ICCV}}.
\newblock


\bibitem[Wen et~al\mbox{.}(2023)]%
        {wen2023treerings}
\bibfield{author}{\bibinfo{person}{Yuxin Wen}, \bibinfo{person}{John
  Kirchenbauer}, \bibinfo{person}{Jonas Geiping}, {and} \bibinfo{person}{Tom
  Goldstein}.} \bibinfo{year}{2023}\natexlab{}.
\newblock \showarticletitle{Tree-Rings Watermarks: Invisible Fingerprints for
  Diffusion Images}. In \bibinfo{booktitle}{\emph{Proc. of {NeurIPS}}}.
\newblock


\bibitem[Wilson(2024)]%
        {fakeart-indi}
\bibfield{author}{\bibinfo{person}{Cam Wilson}.}
  \bibinfo{year}{2024}\natexlab{}.
\newblock \bibinfo{title}{AI is producing 'fake' Indigenous art trained on real
  artists' work without permission}.
\newblock \bibinfo{howpublished}{Crickey.com.au}.
\newblock


\bibitem[YentaMagenta(2023)]%
        {reddithive}
\bibfield{author}{\bibinfo{person}{YentaMagenta}.}
  \bibinfo{year}{2023}\natexlab{}.
\newblock \bibinfo{title}{Hive AI image "detection" is inaccurate and easily
  defeated}.
\newblock \bibinfo{howpublished}{Reddit}.
\newblock


\bibitem[Zhang et~al\mbox{.}(2024)]%
        {zhang2024robust}
\bibfield{author}{\bibinfo{person}{Lijun Zhang} {et~al\mbox{.}}}
  \bibinfo{year}{2024}\natexlab{}.
\newblock \showarticletitle{Robust Image Watermarking using Stable Diffusion}.
\newblock \bibinfo{journal}{\emph{arXiv preprint:2401.04247}}
  (\bibinfo{year}{2024}).
\newblock


\bibitem[Zhu et~al\mbox{.}(2023)]%
        {zhu2023genimage}
\bibfield{author}{\bibinfo{person}{Mingjian Zhu} {et~al\mbox{.}}}
  \bibinfo{year}{2023}\natexlab{}.
\newblock \showarticletitle{GenImage: A Million-Scale Benchmark for Detecting
  AI-Generated Image}.
\newblock \bibinfo{journal}{\emph{arXiv preprint:2306.08571}}
  (\bibinfo{year}{2023}).
\newblock


\end{thebibliography}


\section*{Appendix}
\label{sec:appendix}

\begin{figure}[b!]
    \centering 
    \includegraphics[width=0.45\textwidth]{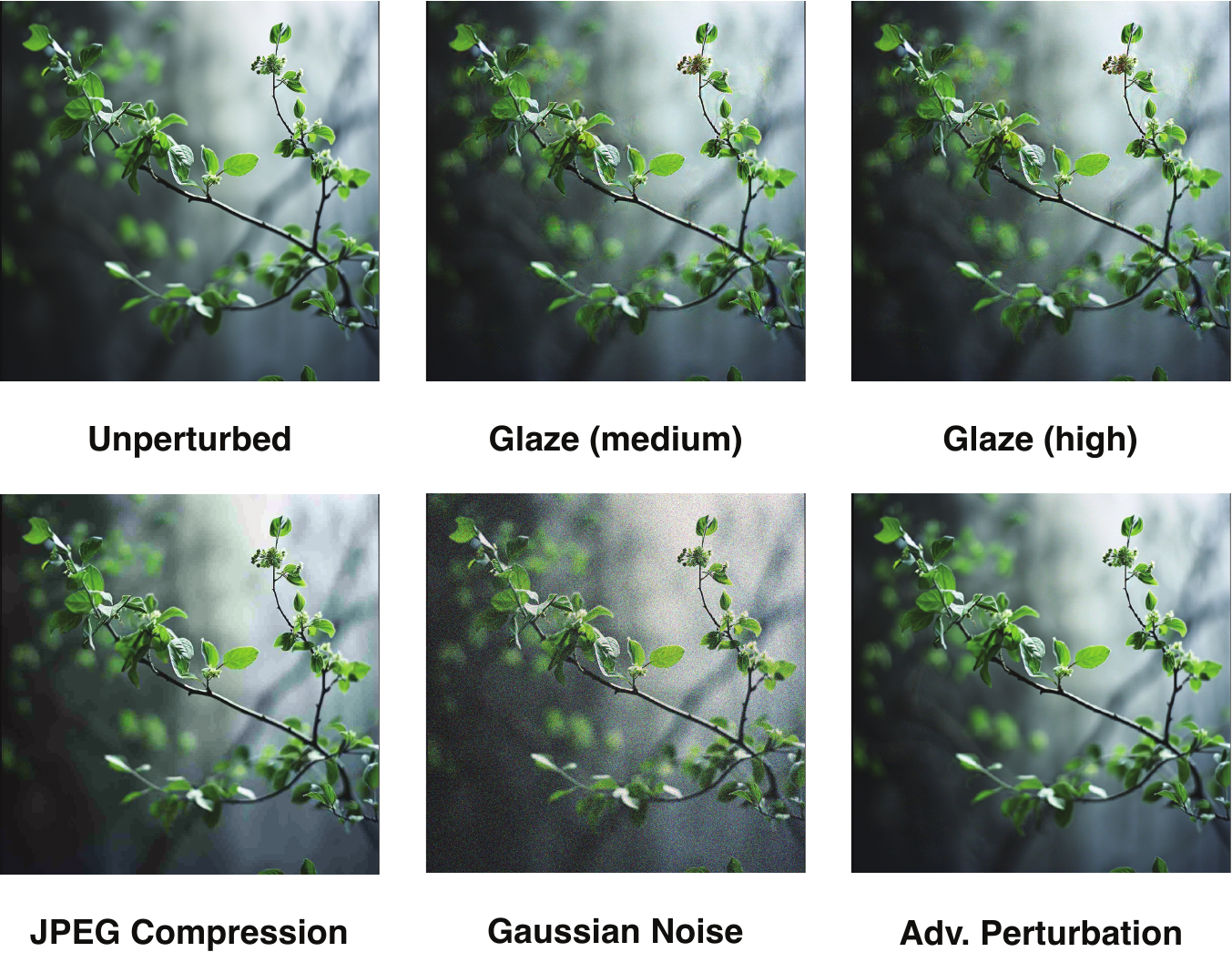}
    \caption{Samples of five different perturbations considered by our study.}
    \label{fig:perturbation-grid2}
\end{figure}

\para{Additional Information on Data Collection.}
\label{app:dataset}
Table~\ref{prompts-modifed} lists the modifications made to the
extracted BLIP captions, which are then used to prompt individual AI
generators.  Figure~\ref{fig:perturbation-grid2} shows 
examples of five types of perturbations considered in  our study.

\para{Detection Results of  DIRE Models.}
Table~\ref{tab:dire-checkpt} shows the detection results using 6 DIRE
checkpoints, using on our dataset of human
artwork and AI-generated images. We test each model with the same set
of 630 unperturbed images: 280 human artworks and 350 AI-generated images.

\para{Additional Results on Hive.} Table~\ref{tab:hive-robust-table}
shows the distribution of images across art style and generative model, which are
``easy-to-detect'' by Hive and unaffected by all five perturbations. 

\begin{figure}[b!]
    \centering
    \includegraphics[width=0.42\textwidth]{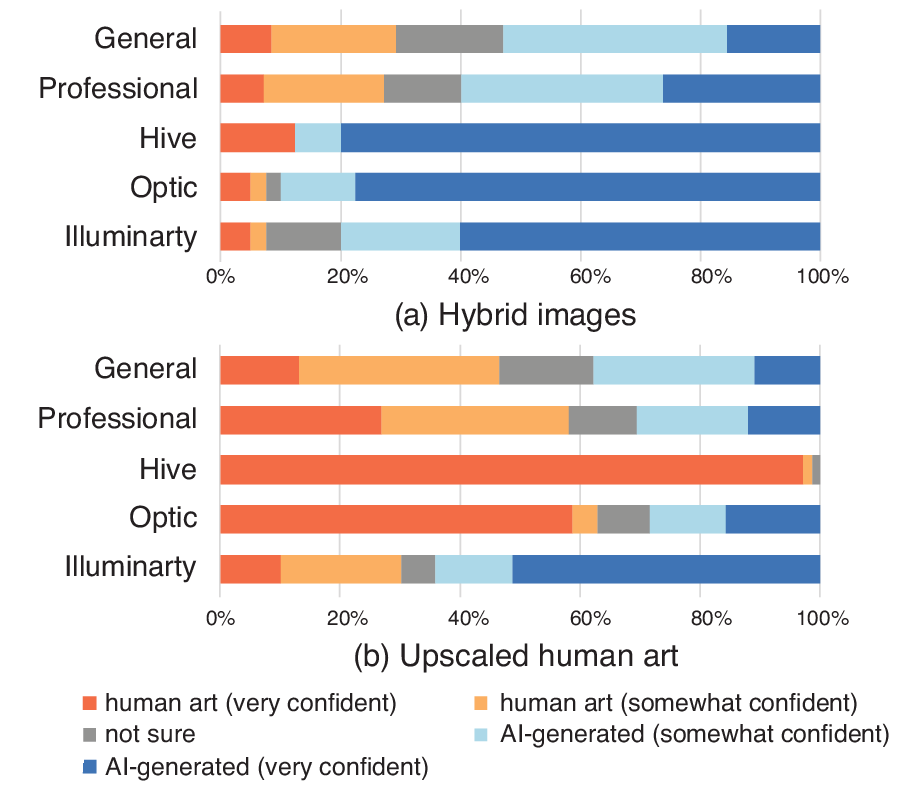}\vspace{-0.1in}
    \caption{Distribution of detection decision represented by the 5-point Likert rating on (a) hybrid images and (b) upscaled human artworks.}
    \label{fig:unusual_decision_confidence}
\end{figure}

\para{Results on Unusual Images.}
Figure~\ref{fig:unusual_decision_confidence} plots, for both hybrid
images and upscaled human art, the distribution of the ``raw''
detection decision represented by the 5-point Likert score. 

\begin{table}[b!]
    \centering
    \resizebox{0.3\textwidth}{!}{
    \centering
    \begin{tabular}{cc}
    \multicolumn{2}{c}{Modified Prompt for Each Style}                                                             \\ \hline
    \multicolumn{1}{c|}{Style}       & \begin{tabular}[c]{@{}c@{}}Modified Prompt:\\ added the phrase\end{tabular} \\ \hline
    \multicolumn{1}{c|}{Anime}       & "anime"                                                                     \\
    \multicolumn{1}{c|}{Cartoon}     & a cartoon style image of"                                                   \\
    \multicolumn{1}{c|}{Fantasy}     & "a fantasy style image of"                                                  \\
    \multicolumn{1}{c|}{Oil/Acrylic} & "an oil and acrylic painting of"                                            \\
    \multicolumn{1}{c|}{Photography} & "a photography of"                                                          \\
    \multicolumn{1}{c|}{Sketch}      & "a sketch drawing of"                                                       \\
    \multicolumn{1}{c|}{Watercolor}  & "a watercolor painting of"                                                 \\ \hline
    \end{tabular}}
    \caption{Phrases added to the BLIP caption extracted from human
      arts to recover the misplaced art
      styles. These captions are then used to prompt AI generators. }
    \label{prompts-modifed}
    \end{table}

\begin{table}[b!]
    \centering
    \resizebox{0.4\textwidth}{!}{
    \centering
\begin{tabular}{cc|ccc}
\multicolumn{2}{c|}{DIRE model}                                                                                          & \multirow{2}{*}{ACC(\%)$\uparrow$} & \multirow{2}{*}{FPR(\%)$\downarrow$} & \multirow{2}{*}{FNR(\%)$\downarrow$}  \\ \cline{1-2}
\begin{tabular}[c]{@{}c@{}}Training\\ dataset\end{tabular} & \begin{tabular}[c]{@{}c@{}}Generation \\ model\end{tabular} &                          &                          &                           \\ \hline
ImageNet                                                   & ADM                                                         & 54.44                    & 96.43                    & 4.86                    \\
LSUN                                                       & PNDM                                                        & 54.76                    & 100.00                   & 1.43    \\
LSUN                                                       & StyleGAN                                                    & 55.40                    & 99.29                    & 0.86      \\
LSUN                                                       & ADM                                                         & 55.08                    & 99.29                    & 1.42      \\
LSUN                                                       & iDDPM                                                       & 54.45                    & 97.86                    & 3.71         \\
CelebA-HQ                                                  & SD-v2                                                       & 51.59                    & 25.36                    & 66.86   \\\hline
\end{tabular}}
\caption{Performance of six DIRE checkpoint models on our dataset. }
\label{tab:dire-checkpt}
\vspace{-0.2in}
\end{table}

\begin{table}[b!]
  \centering
  \resizebox{0.42\textwidth}{!}{
  \centering
  \begin{tabular}{c|ccccc|c}
              & CIVITAI & DALL-E 3 & Firefly & MJv6 & SDXL & Total \\ \hline
  Anime       & 4       & 4        & 9       & 2            & 6    & 25    \\
  Cartoon     & 5       & 6        & 8       & 1            & 6    & 26    \\
  Fantasy     & 10      & 4        & 5       & 0            & 6    & 25    \\
  Oil/Acrylic & 6       & 6        & 1       & 0            & 1    & 14    \\
  Photography & 1       & 3        & 0       & 0            & 2    & 6     \\
  Sketch      & 2       & 1        & 1       & 2            & 3    & 9     \\
  Watercolor  & 5       & 5        & 7       & 1            & 6    & 24    \\ \hline
  Total       & 33      & 29       & 31      & 6            & 30   & 129  \\ \hline
  \end{tabular}}
  \caption{The number of AI-generated images such that for all perturbations, 
  Hive is at least 99\% confident the images are AI-generated. 
  Each column is a generative model, 
  and each row is a style.}  
\label{tab:hive-robust-table}
\end{table}

\end{document}